\documentclass{article}

\usepackage{arxiv}
\usepackage{xcolor}
\usepackage[utf8]{inputenc} 
\usepackage[T1]{fontenc}    
\usepackage[colorlinks,urlcolor=blue]{hyperref}       
\usepackage{url}            
\usepackage{booktabs}       
\usepackage{amsfonts}       
\usepackage{nicefrac}       
\usepackage{microtype}      
\usepackage{lipsum}		
\usepackage{graphicx}
\usepackage{caption}
\usepackage{subcaption}
\usepackage{doi}
\usepackage{amsmath}

\usepackage[ruled,vlined]{algorithm2e}

\title{Attack-SAM: Towards Attacking Segment Anything Model  With Adversarial Examples}

\author{Chenshuang Zhang \\
	KAIST\\
 \And
Chaoning Zhang\thanks{Correspondence Author: chaoningzhang1990@gmail.com} \\
	Kyung Hee University\\
 \And
Taegoo Kang \\
	Kyung Hee University\\
 \And
Donghun Kim \\ 
	Kyung Hee University\\
 \And
Sung-Ho Bae \\
	Kyung Hee University\\
\And
In So Kweon \\
	KAIST\\
}

\renewcommand{\headeright}{}
\renewcommand{\undertitle}{}
\renewcommand{\shorttitle}{Attack-SAM: Towards Attacking Segment Anything Model  With Adversarial Examples}

\begin{document}
\maketitle

\begin{abstract}

  Segment Anything Model (SAM)  has attracted significant attention recently, due to  its impressive performance on various downstream tasks in a zero-short manner. Computer vision (CV)  area might follow the natural language processing (NLP)  area to embark on a path from task-specific vision models toward foundation models. However, deep vision models are widely recognized as vulnerable to adversarial examples, which fool the model to make wrong predictions with imperceptible perturbation. Such vulnerability to adversarial attacks causes serious concerns when applying deep models to security-sensitive applications. Therefore, it is critical to know whether the vision foundation model SAM can also be fooled by adversarial attacks. To the best of our knowledge, our work is the first of its kind to conduct a comprehensive investigation on how to attack SAM with adversarial examples. With the basic attack goal set to mask removal, we investigate the adversarial robustness of SAM in the full white-box setting and transfer-based black-box settings. Beyond the basic goal of mask removal, we further investigate and find that it is possible to generate any desired mask by the adversarial attack.
\end{abstract}

\section{Introduction}

Foundation models~\cite{bommasani2021opportunities} have made significant breakthroughs in the NLP area, with the impressive zero-shot performance of large language models~\cite{brown2020language,ouyang2022training} on various downstream tasks. As a representative large language model,  ChatGPT~\cite{zhang2023ChatGPT} has revolutionized the perceptions of AI and  also falls into the field of generative AI~\cite{zhang2023complete,zhang2023graph_survey} for its ability to generate text. The success of large language models also accelerates the development of multi-modality tasks, such as text-to-image~\cite{zhang2023text} and  text-to-speech~\cite{zhang2023audio}. By contrast, the progress of foundation models for computer vision has somewhat lagged behind. A widely known way to gain a vision foundation model is through contrastive language image pretraining~\cite{jia2021scaling,yuan2021florence}, such as CLIP~\cite{radford2021learning}. Without resorting to language, masked autoencoder~\cite{zhang2022survey} is another popular approach for obtaining foundation models. However, those foundation models often need to be finetuned on the target dataset before they can be used for the downstream tasks, which hinders their generic nature. Following NLP to exploit prompt engineering for adapting the model to downstream tasks without finetuning, very recently, Segment Anything Model (SAM) ~\cite{kirillov2023segment} has proposed a task termed promptable segmentation to train a model to cut out objects on the images in the form of masks. Despite impressive performance in zero-shot transfer tasks, it is unclear whether SAM can be robust against adversarial attack and our work is the first to investigate how to attack SAM with adversarial examples. 

We follow the common practices of popular adversarial attack methods like FGSM attack~\cite{goodfellow2014explaining} and PGD attack~\cite{madry2017towards}. As the first foundation model to perform promptable segmentation, SAM differs from traditional image recognition (segmentation)  models in two ways: (1)  it outputs a mask without label prediction and (2)  it relies on prompt inputs (like points or boxes) . In other words, existing adversarial attacks mainly focus on how to manipulate their image-level (or pixel-level)  label prediction without prompts. To this end, we propose a framework termed Attack-SAM, which is formulated for attacking SAM on the prompt-based mask prediction task. Specifically, we propose a mask removal problem as the basic attack, which is guided by a simple yet effective loss termed ClipMSE. Experimental results show that adversarial attack can successfully remove the predicted mask, suggesting SAM is vulnerable to adversarial examples. We further experiment with transfer-based attacks in two setups: cross-prompt transfer and cross-task transfer. The success of cross-prompt transfer suggests that attackers do not need to know the given point prompt  to guarantee a successful attack. With cross-task transfer, we find that adversarial examples generated with models for semantic label prediction can realize partial success of attacking SAM for prompt-based mask prediction.

Beyond the basic goal of mask removal, we further investigate whether adversarial examples can attack SAM to generate any desired mask. To this end, we design the desired task in three setups by setting it to (1)  a manually designed mask at a random position,  (2)  a mask generated on the same image under a different point prompt, (3)  a mask generated on another image with a random point prompt. Overall, we find it is possible to generate the desired mask properly in most cases, which further highlights the adversarial vulnerability of SAM. Overall, the contributions of this work are summarized as follows.

\begin{itemize}
    \item We conduct the first yet comprehensive investigation on attacking SAM with adversarial examples. We propose a framework for attacking SAM by setting the attack goal as mask removal and design a simple yet effective ClipMSE loss to realize the goal. 
    \item We reveal that SAM is vulnerable to adversarial attacks in full white-box setting. Moreover, we demonstrate that the adversary can attack the model without knowing the prompt and that the SAM can be partially attacked by adversarial examples in a cross-task setup.
    \item Beyond the basic attack goal of mask removal, we further investigate and successfully show that the SAM can be attacked to generate any desired mask, which further highlights the vulnerability of SAM. 
\end{itemize}

The rest of this work is organized as follows. Section~\ref{sec:related} introduces related work by summarizing the progress of SAM and various attack methods. Section~\ref{sec:mask_removal_framework} proposes the framework for Attack-SAM by formulating  mask removal as the attack goal. Section~\ref{sec:mask_removal_results} reports the results of Attack-SAM in the white-box setting. Section~\ref{sec:transfer} reports our investigation of transfer-based attacks on SAM. In Section~\ref{sec:beyond}, we further experiment with how to attack SAM to generate any desired mask. Section~\ref{sec:discussion} discusses the relationship between attacking label prediction and attacking mask prediction, and the limitation of our work.

\section{Related work} \label{sec:related}
\subsection{Segment Anything Model (SAM) } Within less than one month after the advent of SAM, numerous projects and papers have investigated it from various angles, which can be roughly divided into the following categories. A mainstream line of works has tested whether SAM can really segment anything in real-world scenarios, such as medical images~\cite{ma2023segment,zhang2023input}, Camouflaged Object~\cite{tang2023can} and glass (transparent object and mirror) ~\cite{han2023segment}. Their findings show that SAM often fails to detect objects in those challenging scenarios. A similar finding has also been found in~\cite{chen2023sam}, which proposed to use a task-specific adapter to improve SAM in those challenging scenarios. SNA~\cite{jing2023segment} is a pioneering work to extend the SAM to Non-Euclidean Domains with multiple preliminary solutions proposed. Another mainstream line of work has attempted to augment SAM for improving its utility. For example, Grounded SAM~\cite{GroundedSegmentAnything2023} realizes detecting and segmenting anything with text inputs via combing Grounding DINO~\cite{liu2023grounding} with SAM. Given that the generated masks of SAM have no label predictions, multiple works~\cite{chen2023semantic,park2023segment} have attempted to combine SAM with BLIP~\cite{li2022blip} or CLIP~\cite{radford2021learning}. Some works have also utilized SAM for image editing~\cite{kevmo314_2023,feizc_2023,rombach2022high} and inpainting~\cite{yu2023inpaint}. Moreover, SAM has been applied in~\cite{yang2023track,z-x-yang_2023} for tracking objects in video and in~\cite{shen2023anything,kang2022any} for reconstructing 3D objects from a single image.

\subsection{Adversarial attacks}

Deep neural networks are widely known to be vulnerable to adversarial examples, including CNN~\cite{szegedy2013intriguing,goodfellow2014explaining,kurakin2016adversarial} and vision transformer(ViT) ~\cite{dosovitskiy2020image,benz2021adversarial,bhojanapalli2021understanding,mahmood2021robustness}. This vulnerability  has inspired numerous works to investigate the model robustness under various types of adversarial attacks. Adversarial attack methods can be divided into white-box setting~\cite{goodfellow2014explaining,carlini2017towards,madry2017towards} which allows full access to the target model and black-box attacks~\cite{dong2018boosting,xie2019improving,liu2016delving,tramer2017ensemble,wu2020skip,guo2020backpropagating,zhang2022investigating} that mainly rely on transferability of adversarial examples. Another way to categorize the attacks is untargeted and targeted. In the context of image recognition (classification), an attack is considered as successful if the predicted label is different from the ground-truth label under the untargeted setting. In the more strict targeted setting, the attack fails if the predicted label is not the pre-determined target label. The above works have mainly focused on manipulating the image-level label prediction for image classification tasks, while our work investigates how to attack SAM for the task of prompt-based mask prediction. Since SAM is the first work to achieve prompt-based mask prediction, it remains unclear whether SAM can be robust against adversarial attacks. It is worth mentioning that attacking SAM is also different from attacking semantic segmentation models,  since the generated masks have no semantic labels. 

\section{Framework for Attack-SAM}
\label{sec:mask_removal_framework}

\subsection{Preliminaries}
\label{sec:preliminaries}

\textbf{Prompotable mask prediction.} The task  SAM solves is promptable segmention,  where the model generates masks as the outputs by taking both images and prompts as the inputs. It is worth noting that the generated masks just cut out the detected objects but has no semantic labels for each mask. Therefore, the data pair for SAM could be represented by ($x$, $prompt$, $Mask$), where $prompt$ is necessary for each prediction.
For a certain image $x$, the prompt could be selected randomly across the image, leading to multiple data pairs  $\mathbb{D} = \{$($x$, $prompt_{i}$, $Mask_{i}$) $\}$.

The forward progress of SAM is given as follows: 
\begin{equation}
y = SAM(promt, x; \theta) 
\label{eq:sam_forward}
\end{equation}
where $y$ indicates the confidence of being masked for each pixel. In the vanilla setting of SAM, $y$ has the shape of $H*W$ where H, W indicate the height and width of input image, respectively. With $i$ and $j$ indicate the coordinates of pixel in the image, the pixel $x_{ij}$ is marked within the mask area if the predicted value $y_{ij}$ for $x_{ij}$ is positive (larger than zero). Otherwise, it is marked as the background. The final predicted masks are termed as $Mask_{pred}$, which is a binary matrix in the shape of $H*W$ .

\textbf{Common attack methods.}
Before introducing Attack-SAM, we first revisit the widely applied attack methods in the traditional classification task. We define $f(\cdot ,  \theta) $ as the to-be-attacked target model, parameterized by $\theta$. With ($x_{clean}$, $Y$)  as data pairs from the original dataset, the adversarial image $x_{adv}$ is defined as $x+\delta^{*}$, where $\delta^{*}$ is optimized in Equation~\ref{eq:attack}. Specifically, the attack algorithm is designed to generate the optimal  $\delta^{*}$. In the classification task, $Y$ often indicates the class label and $loss$ is often cross-entropy function. 

\begin{equation}
 \delta^{*} = \max_{\delta \in \mathbb{S}} loss(f(x_{clean} + \delta; \theta) , Y) 
\label{eq:attack}
\end{equation}

Fast Gradient Sign Method (FGSM) ~\cite{goodfellow2014explaining} and Projected Gradient Descent (PGD) ~\cite{madry2017towards}  are two widely used methods for evaluating the model robustness for their simplicity and effectiveness. FGSM is a single-step attack method based on the model gradient on the input image. PGD can be seen as an iterative version of FGSM, which is also termed I-FGSM in some works. In this work, we stick to term it PGD for consistency. If the iteration number is N, the PGD attack is denoted as PGD-N.

\subsection{Attack-SAM}
\label{sec:definition_mask_removal}

\textbf{Task definition.} For typical adversarial attacks on image recognition models, the goal is to change the image-wise or pixel-wise predicted labels so that the model makes wrong predictions. Given that the generated masks from SAM have no semantic labels, a straightforward way to successfully attack SAM is to make the model fail to detect the objects so that the generated masks are removed after adding adversarial perturbation. In this work, we perceive mask removal as the basic goal of adversarial attacks on SAM. According to Section~\ref{sec:preliminaries}, a pixel $x_{ij}$ is predicted as masked if the predicted value $y_{ij}$ is positive. Therefore, the mask removal task is successful if the predicted values $y$ in the target area all become negative. In other words, we can attack SAM to remove masks by reducing $y$ until being negative.

\textbf{Loss design.} To remove masks by attacking SAM, the loss design is expected to be made to reduce the predicted values $y$ until being negative. Moreover, for positive $y_{ij}$ with large amplitude, the loss is expected to punish them more than the positive $y_{mn}$ with small amplitude or already negative ones. To reduce the randomness effect, it is desired to make the predicted values lower than a certain negative value instead of being slightly lower than zero.  With these three expectations, the Mean Squared Error  (MSE)  loss with a negative threshold is a natural choice. As shown in Equation~\ref{eq:mse_loss}, the predicted value $SAM(prompt, x_{clean} + \delta) $  is optimized to be close to a negative threshold $Neg_{th}$ after attack. In the extreme case of $SAM(prompt, x_{clean} + \delta) =Neg_{th}$ for all predicted values $y$, the MSE loss achieves its minimum: zero.

\begin{equation}
  \delta^{*}  = \min_{\delta \in \mathbb{S}}  ||SAM(prompt, x_{clean} + \delta) - Neg_{th}||^2
\label{eq:mse_loss}
\end{equation}

The above MSE loss in Equation~\ref{eq:mse_loss} could optimize $\delta$ to predict negative values and also meets the expectations discussed above. For the predicted values already lower than $Neg_{th}$, however, it makes no sense to increase them to be close to  $Neg_{th}$. To mitigate this issue, we clip the predicted values smaller than $Neg_{th}$ so that the loss leads to zero gradients on those pixel values. We term this loss as ClipMSE, as shown in Equation~\ref{eq:clipmse_loss}. The superiority of ClipMSE over the vanilla MSE is verified in Table~\ref{tab:remove_single_mask_whitebox}. 

\begin{equation}
\delta^{*}  =  \min_{\delta \in \mathbb{S}}  || Clip(SAM(prompt, x_{clean}+ \delta) ,min=Neg_{th})  - Neg_{th}||^2
\label{eq:clipmse_loss}
\end{equation}

\textbf{Attack details.}  We adopt FGSM~\cite{goodfellow2014explaining} attack and PGD attack~\cite{madry2017towards} which are widely used for robustness evaluation in prior works. Following prior works on attacking vision models in the white-box setting, we set the maximum allowable perturbation magnitude to $8/255$ by default, with a step size of $8/255$ and $2/255$ for FGSM and PGD attack, respectively. If not specified, PGD-10 attack is adopted, which indicates that the iteration number is 10. The threshold $Neg_{th}$ is set to $-10$ in this work, since the predicted values in the background (non-mask)  region are often around this value. 

\section{Main results of Attack-SAM}
\label{sec:mask_removal_results}

In the vanilla version of SAM, a single point is chosen in the image and a single mask near the given point is then predicted. As a basic setup, we first attack SAM to remove a single mask
as discussed in Section~\ref{sec:definition_mask_removal}. Since each mask is generated from an input prompt, the attack succeeds if SAM fails to predict the original mask based on that prompt. Specifically, the attack succeeds in the mask removal task if $Mask_{adv}$ is  empty or at least with a much smaller area than $Mask_{clean}$.  To not lose generality, the ($prompt$, $Mask_{clean}$)  pair to be attacked is selected randomly from the image.

\subsection{Qualitative results}
For visualization examples, we randomly select clean images either from demo images of SAM project or SA-1B dataset of SAM paper~\cite{kirillov2023segment}. The visualization results  of adversarial images and predicted masks are reported in Figure~\ref{fig:remove_single_mask_whitebox}, and more results are given in the appendix.  With a randomly selected point as the prompt (marked as the green star in Figure~\ref{fig:remove_single_mask_whitebox}(a)), the model is able to generate adversarial images with imperceptible perturbations after FGSM and PGD attack (see Figure~\ref{fig:remove_single_mask_whitebox}(b)  and Figure~\ref{fig:remove_single_mask_whitebox}(c)). Although SAM is able to predict a high-quality $Mask_{clean}$ in Figure~\ref{fig:remove_single_mask_whitebox}(d), both FGSM  and PGD attack are able to remove the area of $Mask_{clean}$, especially  the tiny white area of  $Mask_{pgd}$ in Figure~\ref{fig:remove_single_mask_whitebox}(f). Figure~\ref{fig:remove_single_mask_whitebox} shows that SAM is vulnerable to adversarial attacks in the mask removal task, and the PGD attack performs better than the FGSM attack.

\begin{figure*}[!htbp]
     \centering
         \begin{minipage}[b]{0.16\textwidth}
         \includegraphics[width=\textwidth]{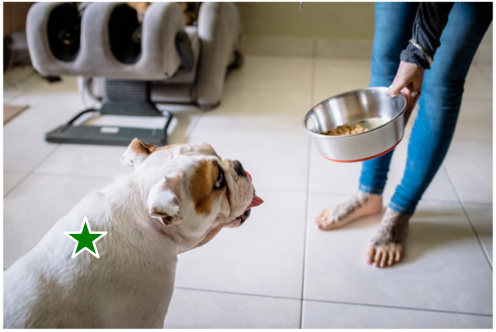}
     \end{minipage}
    \begin{minipage}[b]{0.16\textwidth}
         \includegraphics[width=\textwidth]{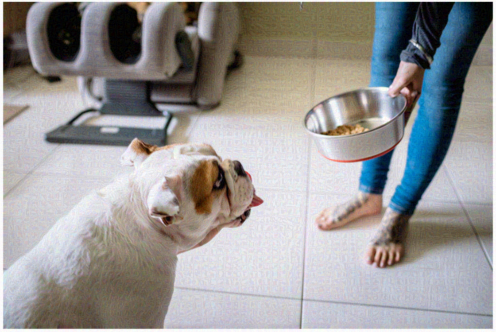}
     \end{minipage}
         \begin{minipage}[b]{0.16\textwidth}
         \centering
         \includegraphics[width=\textwidth]{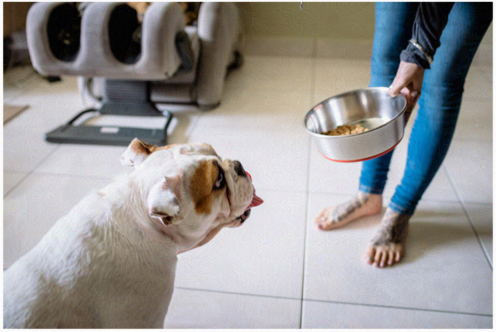}
     \end{minipage}
        \begin{minipage}[b]{0.16\textwidth}
         \centering
         \includegraphics[width=\textwidth]{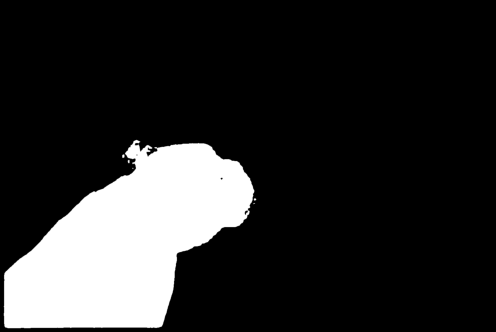}
     \end{minipage}
    \begin{minipage}[b]{0.16\textwidth}
         \centering
         \includegraphics[width=\textwidth]{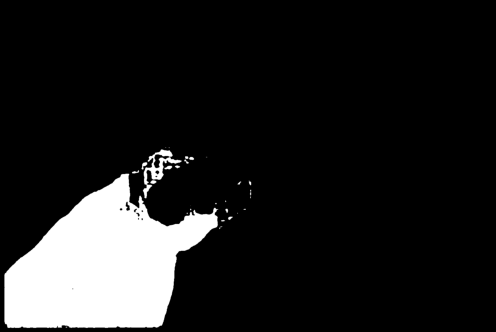}
     \end{minipage}
     \begin{minipage}[b]{0.16\textwidth}
         \centering
         \includegraphics[width=\textwidth]{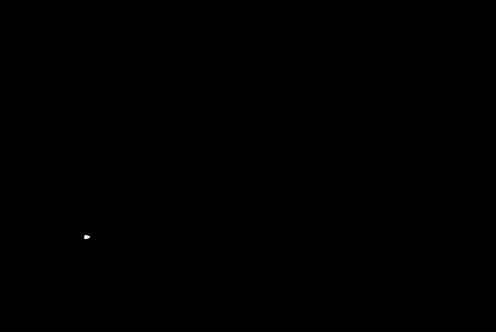}
     \end{minipage}

        \begin{minipage}[t]{0.16\textwidth}
         \includegraphics[width=\textwidth]{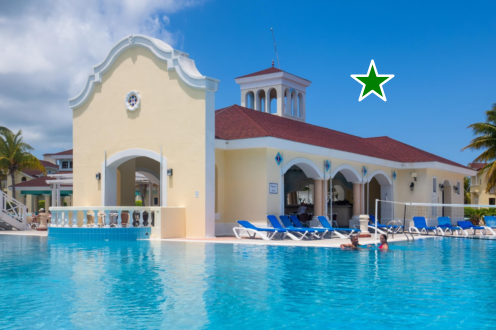}
         \subcaption{$x_{clean}$}
     \end{minipage}
         \begin{minipage}[t]{0.16\textwidth}
         \centering
         \includegraphics[width=\textwidth]{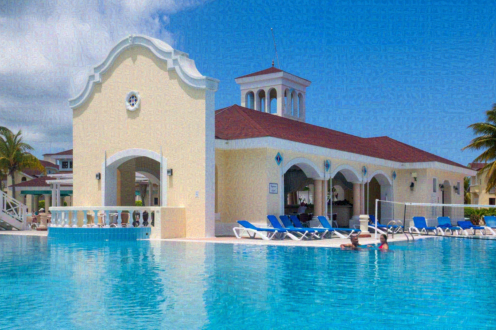}
         \subcaption{$x_{fgsm}$}
     \end{minipage}
     \begin{minipage}[t]{0.16\textwidth}
         \centering
         \includegraphics[width=\textwidth]{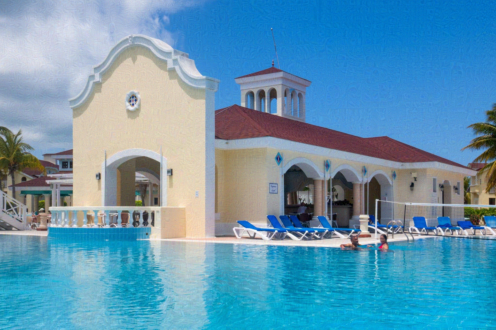}
         \subcaption{$x_{pgd}$}
     \end{minipage}
        \begin{minipage}[t]{0.16\textwidth}
         \centering
         \includegraphics[width=\textwidth]{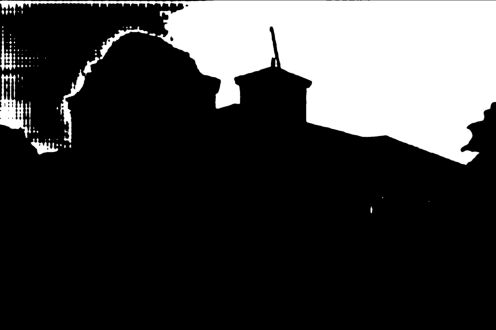}
         \subcaption{$Mask_{clean}$}
     \end{minipage}
     \begin{minipage}[t]{0.16\textwidth}
         \centering
         \includegraphics[width=\textwidth]{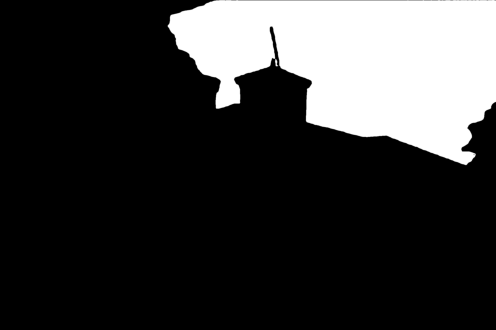}
         \subcaption{$Mask_{fgsm}$}
     \end{minipage}
          \begin{minipage}[t]{0.16\textwidth}
         \centering
         \includegraphics[width=\textwidth]{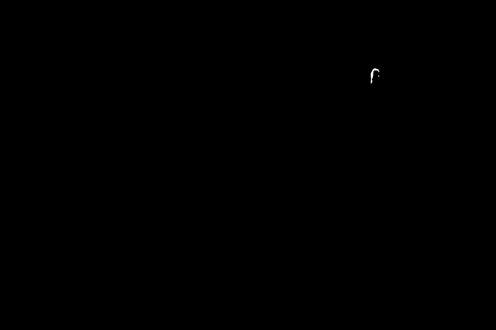}
         \subcaption{$Mask_{pgd}$}
     \end{minipage}
        \caption{Attack SAM to remove the masks. Figure (a)  refers to the clean image with the location of point prompt marked in green star. Figure (b)  and (c)  are adversarial images generated by FGSM and PGD attacks, respectively. The white area in Figure (d) (e) (f)  refer to masks predicted by SAM based on the given prompt and images in Figure (a) (b) (c), respectively. The results in  Figure (e) (f)  show that SAM is vulnerable to adversarial attacks considering the reduce white area compared to Figure (d).
        }
    \label{fig:remove_single_mask_whitebox}
\end{figure*}

\subsection{Quantitative results}

\textbf{Evaluation metric.} For quantitative evaluation of adversarial attack on SAM, we adopt the IoU metric widely used in segmentation. To evaluate the change of masks after attack,  we calculate IoU between the predicted masks of clean image $Mask_{clean}$ and predicted masks of adversarial image $Mask_{adv}$. As shown in Equation~\ref{eq:miou}, an average of IoU from $N$ data pairs is calculated. 
\begin{equation}
mIoU = \frac{1}{N}\sum_{i=1}^{N} IoU(Mask_{adv}, Mask_{clean}) ,
\label{eq:miou}
\end{equation}
It should be noted that $Mask$ in Equation~\ref{eq:miou} is a binary matrix indicating whether a pixel is predicted to be masked.
The maximum of mIoU equals 1 when perturbation $\delta$ is a zero vector, corresponding to the case of no attack. 

We report mIoU results after different attacks in Table ~\ref{tab:remove_single_mask_whitebox}. With the proposed ClipMSE, the mIoU drops from 1 to close to zero after PGD-attack. Although FGSM attack also achieves mIoU much smaller than 1, the result under FGSM attack is worse than PGD attack due to weaker attack strength. This result is consistent with the visualization in Figure~\ref{fig:remove_single_mask_whitebox}. We also compare MSE loss and ClipMSE loss. Although the two losses achieve similar results under the FGSM attack, ClipMSE loss significantly outperforms MSE loss under the PGD attack. This is expected since the advantage of ClipMSE loss is to remove the gradient of pixels with predicted value $y$ smaller than $Neg_{th}$. Meanwhile, it is hard for FGSM attack to make predicted  $y$ smaller than $Neg_{th}$ within a single attack step. Therefore, the advantages of ClipMSE loss  are more significant in multiple-iteration attacks, such as PGD-10.

\begin{table}[!htbp]
  \caption{Results of mIoU in removing mask attack on SAM. Both FGSM and PGD-10 attack achieve much lower mIoU than the setting of no attack, while PGD-10 with ClipMSE achieves the lowest mIoU, outperforming other settings.}
  \label{tab:remove_single_mask_whitebox}
  \centering
  \begin{tabular}{lclllllllllllll}
    \toprule
    Attacks & No Attack  &  \multicolumn{2}{c}{MSE}  & \multicolumn{2}{c}{ClipMSE} \\
    \cline{3-6} 
      &  &FGSM & PGD-10  &FGSM & PGD-10 \\
    \midrule
    mIoU &1.0    &0.4868 & 0.1639 &0.4807 & \textbf{0.0487} \\
  \bottomrule
\end{tabular}
\end{table}

\subsection{Attack SAM under other forms of prompts} 
In this work, we mainly experiment with attacking SAM under the point prompt. However, SAM also accepts other forms of prompts like bounding box. Interested readers can refer to the appendix for the relevant results.

\section{Transfer-based attacks} \label{sec:transfer}
Section~\ref{sec:mask_removal_results} has shown that SAM is vulnerable to adversarial attacks in a full white-box setting. A natural question arises: is SAM robust to transfer-based attacks? In the black-box setting, the attacker cannot access all the information needed when attacking a certain target model. In this section, we evaluate the robustness of SAM by introducing two transfer-based attacks: cross-prompt transfer and cross-task transfer.

\subsection{Cross-prompt transfer}
\label{sec:attack_multiple_masks}

\textbf{Task definition.} In Section~\ref{sec:mask_removal_results}, a single point $prompt_{i}$ is randomly selected as input prompt to generate the adversarial image $x_{adv}$. SAM then fails to predict the corresponding  $Mask_{i}$ with ($prompt_{i}$, $x_{adv}$)  as the input. The question is: could this vulnerability transfer between point prompts? In other words, if we generate $x_{adv}$ by attacking the ($prompt_{i}$, $Mask_{i}$)  as in Section~\ref{sec:mask_removal_results}, could SAM predict valid masks for another point prompt $prompt_{j}$? We term this attack as cross-prompt transfer attack. Specifically, the point prompt  (above $prompt_{i}$)  selected to generate $x_{adv}$ is termed as source prompt ($prompt_{source}$) , and the above $prompt_{j}$ for mask prediction is termed target prompt ($prompt_{target}$) .

\begin{figure}[!htbp]
     \centering
        \begin{minipage}[b]{0.20\textwidth}
         \centering
         \includegraphics[width=\textwidth]{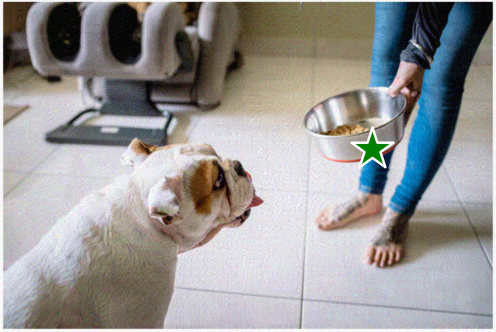}
     \end{minipage}
        \begin{minipage}[b]{0.20\textwidth}
         \centering
         \includegraphics[width=\textwidth]{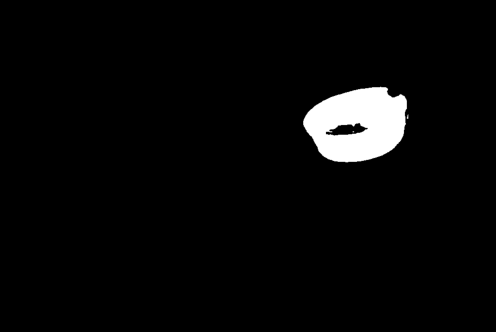}
     \end{minipage}
      \begin{minipage}[b]{0.20\textwidth}
         \centering
         \includegraphics[width=\textwidth]{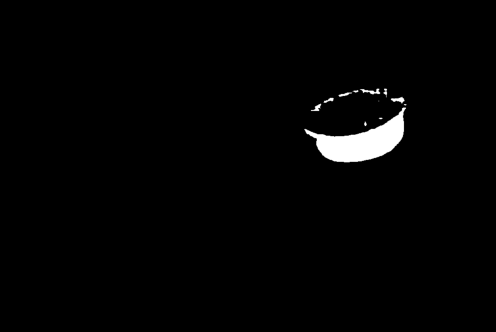}
     \end{minipage}
     
     \begin{minipage}[t]{0.20\textwidth}
         \centering
         \includegraphics[width=\textwidth]{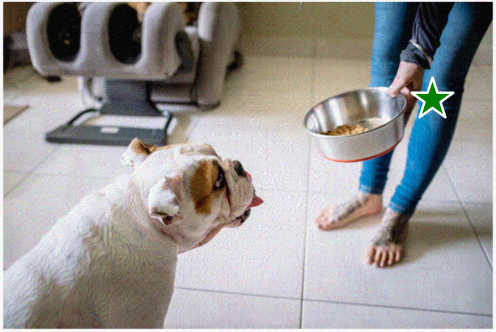}
          \subcaption{Input $x_{pgd}$ with $prompt_{target}$}
     \end{minipage}
        \begin{minipage}[t]{0.20\textwidth}
         \centering
         \includegraphics[width=\textwidth]{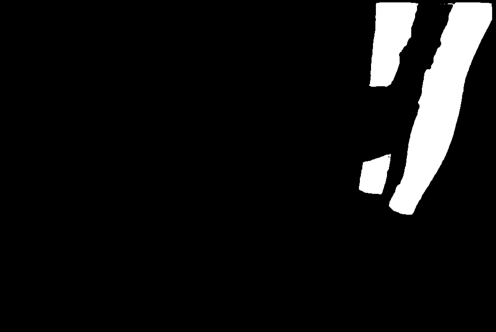}
          \subcaption{Masks predicted by ($x_{clean}$,$prompt_{target}$)}
     \end{minipage}
     \begin{minipage}[t]{0.20\textwidth}
         \centering
         \includegraphics[width=\textwidth]{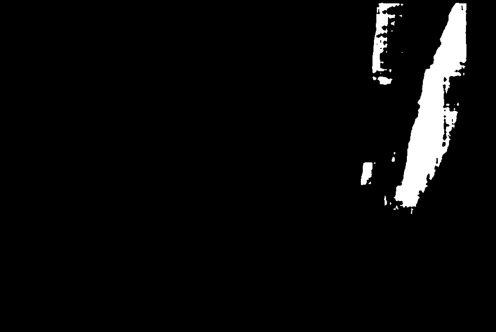}
          \subcaption{Masks predicted by ($x_{pgd}$,$prompt_{target}$)}
     \end{minipage}
        \caption{Preliminary study for corss-prompt transfer. (a)  The adversarial image $x_{pgd}$ is Figure~\ref{fig:remove_single_mask_whitebox}(c)  is adopted for mask prediction with  prompt ($prompt_{target}$)  marked as a green star. Figure (b)  and (c)  are predicted masks based on ($prompt_{target}$, $x_{clean}$)  and ($prompt_{target}$, $x_{pgd}$) , respectively. Note that $prompt_{target}$ in (a)  is different from the prompt ($prompt_{source}$)  when generating $x_{pgd}$ in Figure~\ref{fig:remove_single_mask_whitebox}.
        }
    \label{fig:single_point_transfer}
\end{figure}

\textbf{A preliminary study.} To investigate whether the vulnerability could transfer between different point prompt s, we take the adversarial images $x_{pgd}$ in Figure~\ref{fig:remove_single_mask_whitebox}(c)  as the input image of SAM, and attack SAM to predict masks with prompts ($prompt_{target}$)  different from $prompt_{source}$ marked as the green star in  Figure~\ref{fig:remove_single_mask_whitebox}(a) .
The $x_{pgd}$ is shown in Figure~\ref{fig:single_point_transfer}(a) , with $prompt_{target}$ marked as green star. It should be noted that the $x_{pgd}$ in Figure~\ref{fig:remove_single_mask_whitebox}(a)  is the same image as Figure~\ref{fig:remove_single_mask_whitebox}(c) . As shown in Figure~\ref{fig:single_point_transfer}(c) , SAM could predict masks based on the  ($prompt_{target}$, $x_{pgd}$)  pair in Figure~\ref{fig:single_point_transfer}(a) . However, the quality of generated masks in Figure~\ref{fig:single_point_transfer}(c)  is lower than the masks generated on $x_{clean}$ in Figure~\ref{fig:single_point_transfer}(b) . These observations show that generating $x_{pgd}$ from $prompt_{source}$ can cause vulnerability of other (prompt, mask)  pairs to some extent (see the reduced area of masks in Figure~\ref{fig:single_point_transfer}(c)  compared to Figure~\ref{fig:single_point_transfer}(b) ) . However, there still exist valid masks in Figure~\ref{fig:single_point_transfer}(c)  with $x_{pgd}$ as input. The results in Figure~\ref{fig:single_point_transfer} inspire us to investigate the possible reason for the above phenomenon.

\begin{figure}[!htbp]
     \centering
      \begin{minipage}[b]{0.20\textwidth}
         \centering
         \includegraphics[width=\textwidth]{figs_attack/prompt_transfer_single_point/example1/adv_seg_result.png}
     \end{minipage}
      \begin{minipage}[b]{0.20\textwidth}
         \centering
         \includegraphics[width=\textwidth]{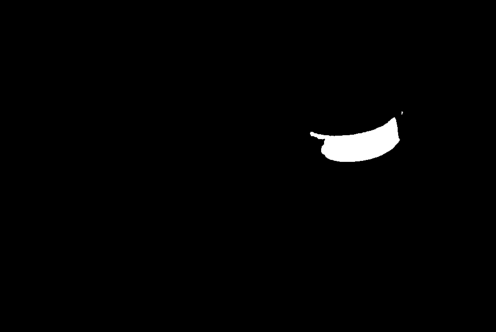}
     \end{minipage}  
      \begin{minipage}[b]{0.20\textwidth}
         \centering
         \includegraphics[width=\textwidth]{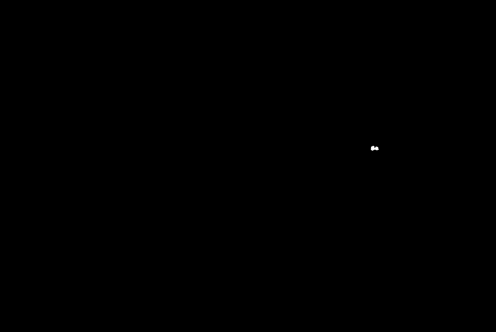}
     \end{minipage}
     
     \begin{minipage}[t]{0.20\textwidth}
         \centering
         \includegraphics[width=\textwidth]{figs_attack/prompt_transfer_single_point/example2/adv_seg_result.png}
         \subcaption{$K$=1 }
     \end{minipage}
      \begin{minipage}[t]{0.20\textwidth}
         \centering
         \includegraphics[width=\textwidth]{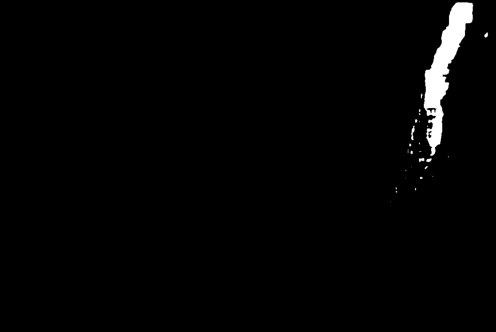}
         \subcaption{$K$=100}
     \end{minipage}  
      \begin{minipage}[t]{0.20\textwidth}
         \centering
         \includegraphics[width=\textwidth]{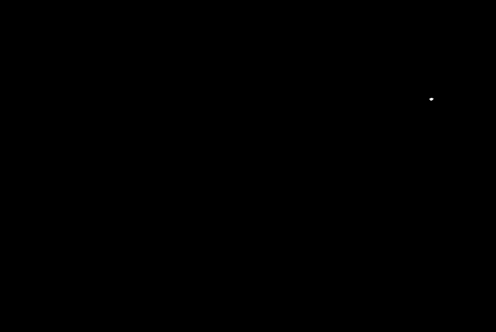}
         \subcaption{$K$=400}
     \end{minipage}
    \caption{Masks predicted on ($x_{pgd}$, $prompt_{target}$)  with different numbers of $prompt_{source}$ ($K$)  when generating adversarial images. The masks in white shrink significantly as $K$ increases.}
    \label{fig:single_point_transfer_multiple_masks}
\end{figure}

\begin{figure*}[!htbp]
     \centering
      \begin{minipage}[b]{0.18\textwidth}
         \centering
         \includegraphics[width=\textwidth]{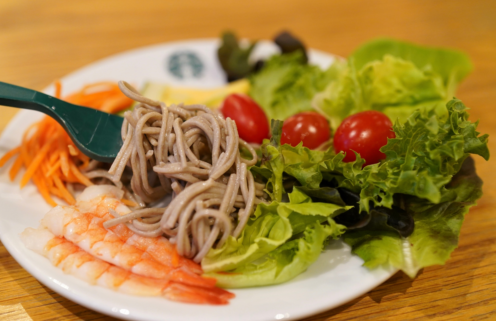}
     \end{minipage}
      \begin{minipage}[b]{0.18\textwidth}
         \centering
         \includegraphics[width=\textwidth]{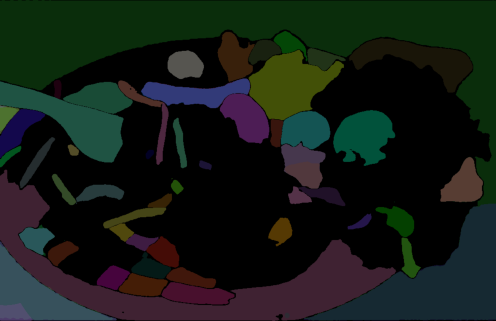}
     \end{minipage}  
      \begin{minipage}[b]{0.18\textwidth}
         \centering
         \includegraphics[width=\textwidth]{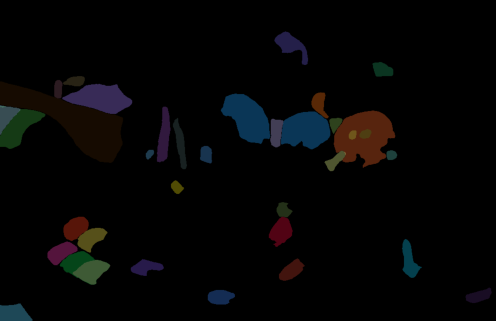}
     \end{minipage}   
      \begin{minipage}[b]{0.18\textwidth}
         \centering
         \includegraphics[width=\textwidth]{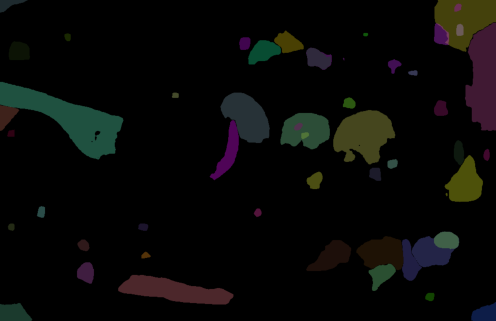}
     \end{minipage}
    \begin{minipage}[b]{0.18\textwidth}
         \centering
         \includegraphics[width=\textwidth]{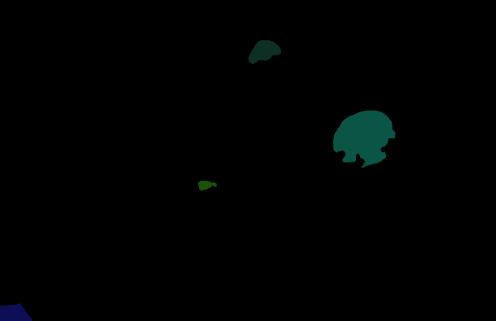}
     \end{minipage}

      \begin{minipage}[b]{0.18\textwidth}
         \centering
         \includegraphics[width=\textwidth]{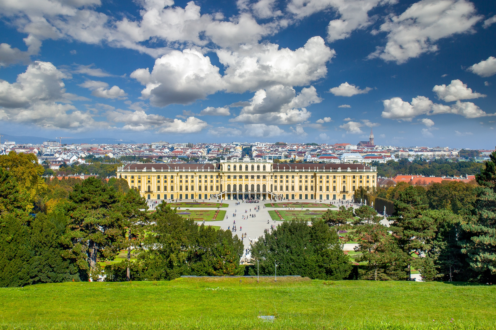}
     \end{minipage}
      \begin{minipage}[b]{0.18\textwidth}
         \centering
         \includegraphics[width=\textwidth]{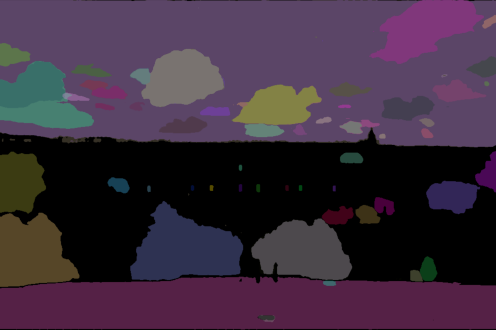}
     \end{minipage}  
      \begin{minipage}[b]{0.18\textwidth}
         \centering
         \includegraphics[width=\textwidth]{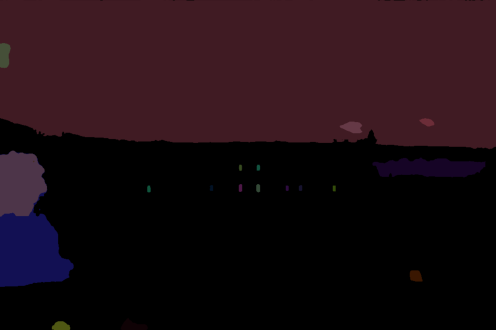}
     \end{minipage}   
      \begin{minipage}[b]{0.18\textwidth}
         \centering
         \includegraphics[width=\textwidth]{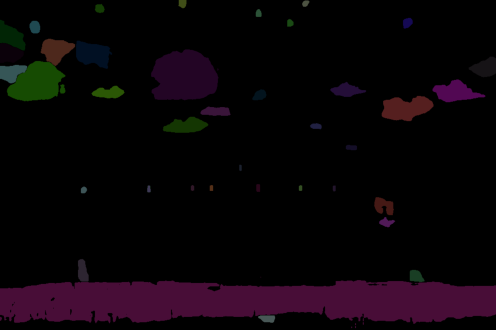}
     \end{minipage}
    \begin{minipage}[b]{0.18\textwidth}
         \centering
         \includegraphics[width=\textwidth]{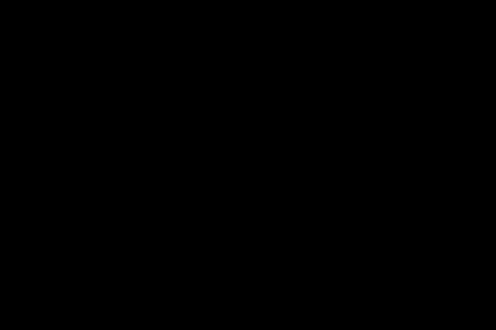}
     \end{minipage}

     \begin{minipage}[t]{0.18\textwidth}
         \centering
         \includegraphics[width=\textwidth]{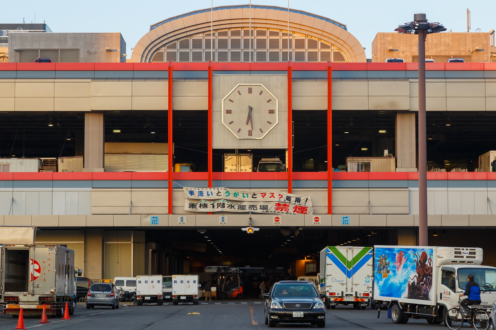}
      \subcaption{$x_{clean}$}
     \end{minipage}
      \begin{minipage}[t]{0.18\textwidth}
         \centering
         \includegraphics[width=\textwidth]{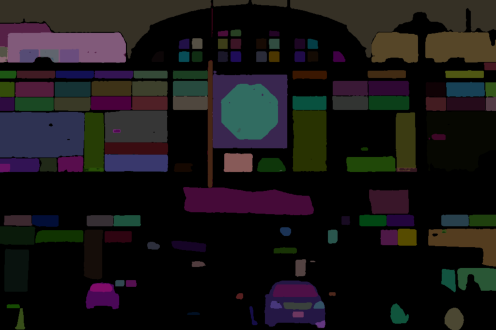}
     \subcaption{$Mask_{clean}$}
     \end{minipage}  
      \begin{minipage}[t]{0.18\textwidth}
         \centering
         \includegraphics[width=\textwidth]{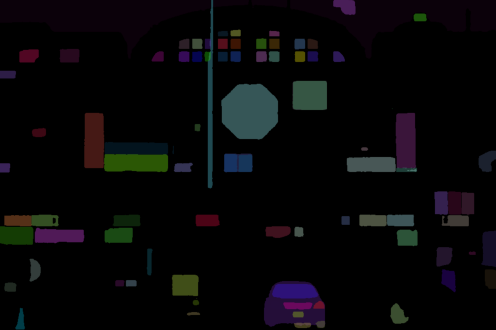}
     \subcaption{$Mask_{adv}$ ($K$=1) }
     \end{minipage}   
      \begin{minipage}[t]{0.18\textwidth}
         \centering
         \includegraphics[width=\textwidth]{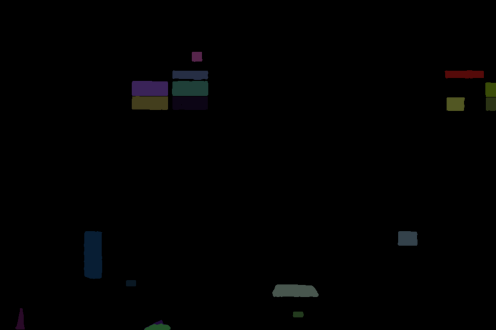}
         \subcaption{$Mask_{adv}$ ($K$=100) }
     \end{minipage}
    \begin{minipage}[t]{0.18\textwidth}
         \centering
         \includegraphics[width=\textwidth]{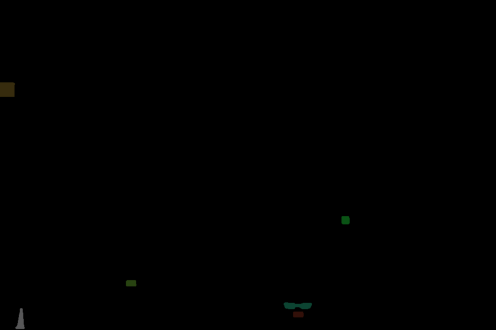}
           \subcaption{$Mask_{adv}$ ($K$=400) }
     \end{minipage}
    \caption{Masks predicted on multiple ($x$, $prompt_{target}$)  pairs in cross-prompt transfer task,  with different numbers of $prompt_{source}$ ($K$)  when generating adversarial images. With each colored region referring to a mask, the valid masks $Mask_{adv}$ reduce significantly as $K$ increases.
    }
    \label{fig:prompt_transfer_multiple_masks}
\end{figure*}

\textbf{Enhancing cross-prompt transferability.} We propose a possible explanation for the interesting phenomenon in Figure~\ref{fig:single_point_transfer} as follows. The reason is that SAM by default generates the mask around the given point prompt . When the generated perturbation in $x_{pgd}$ mainly focuses on removing the mask around the selected $prompt_{source}$, it can be less effective in other regions. To verify this hypothesis, we increase the number ($K$)  of $prompt_{source}$ when generating $x_{pgd}$. The results in Table~\ref{tab:remove_multiple_mask_point_number} and Figure~\ref{fig:single_point_transfer_multiple_masks} show that the mask predicted by ($prompt_{target}$, $x_{pgd}$)  could be removed as $K$ increases. The above example shows the setting when the number of $prompt_{target}$ is 1, and we also report the masks predicted on  multiple ($x$, $prompt_{target}$) pairs in Figure~\ref{fig:prompt_transfer_multiple_masks}. As shown in Figure~\ref{fig:prompt_transfer_multiple_masks}(c), when the number of  $prompt_{source}$ ($K$) is 1, the number of predicted masks is less than that in Figure~\ref{fig:prompt_transfer_multiple_masks}(b), but there are still valid masks remaining in Figure~\ref{fig:prompt_transfer_multiple_masks}(c). As the number of  $prompt_{source}$ ($K$) increases, more masks could be removed in the multiple $prompt_{target}$ setting. This conclusion is consistent with that in Table~\ref{tab:remove_multiple_mask_point_number} and Figure~\ref{fig:single_point_transfer_multiple_masks} where the number of $prompt_{target}$ is 1. These results show that the adversary can attack the model without knowing the prompt, and more masks could be removed as the number of  $prompt_{source}$ increases. More investigations and understandings of the cross-prompt task are encouraged in future works.

\begin{table}[!htbp]
  \caption{Results of cross-prompt transfer with different numbers of $prompt_{source}$ when generating adversarial images, termed $K$. The mIoU decreases significantly as $K$ increases, indicating more successful attack.
  }
  \label{tab:remove_multiple_mask_point_number}
  \centering
  \begin{tabular}{llllllllllllll}
    \toprule
        $K$  &  1  & 100  &  400\\
    \midrule
    mIoU   &  0.4234  & 0.1247 &  0.0971\\
  \bottomrule
\end{tabular}
\end{table}

\subsection{Cross-task transfer}
\label{sec:cross_task_transfer}

\textbf{Task definition.} The task of SAM is to predict masks based on prompts. Therefore, a successful attack can mislead SAM to make wrong decisions for mask prediction. With various deeply explored models on label classification task, we raise the following question: could SAM predict valid masks based on the adversarial images $x_{adv}$ generated on models from other tasks, such as classification? Since the task to generate $x_{adv}$ ($task_{source}$)  is different from the mask prediction task in SAM ($task_{target}$), we term this attack as cross-task transfer attack.

\begin{figure}[!htbp]
     \centering
         \begin{minipage}[b]{0.12\textwidth}
         \includegraphics[width=\textwidth]{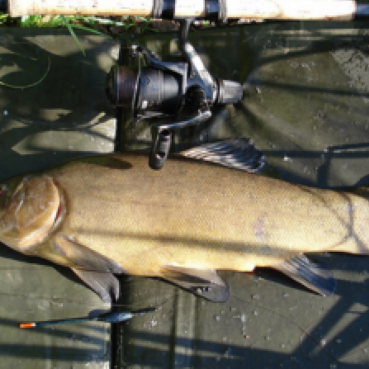}
     \end{minipage}
    \begin{minipage}[b]{0.12\textwidth}
         \includegraphics[width=\textwidth]{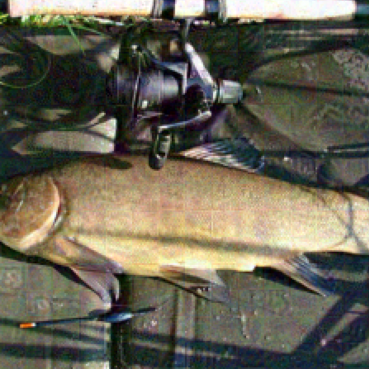}
     \end{minipage}
    \begin{minipage}[b]{0.12\textwidth}
         \centering
         \includegraphics[width=\textwidth]{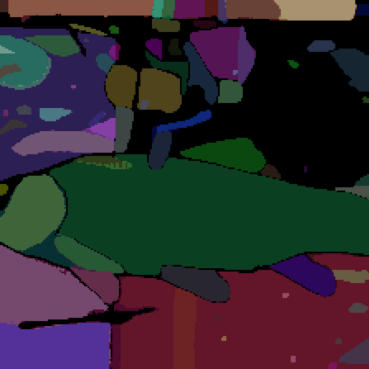}
     \end{minipage}
        \begin{minipage}[b]{0.12\textwidth}
         \centering
         \includegraphics[width=\textwidth]{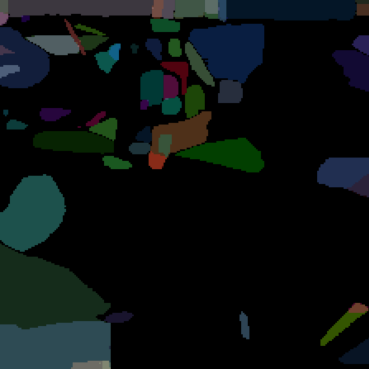}
     \end{minipage}

    \begin{minipage}[b]{0.12\textwidth}
         \includegraphics[width=\textwidth]{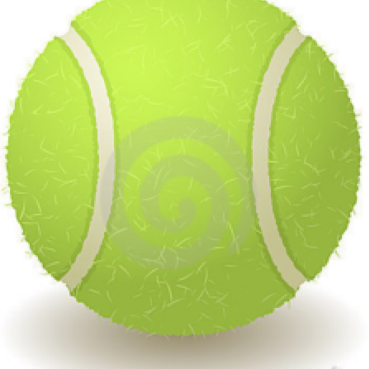}
     \end{minipage}
    \begin{minipage}[b]{0.12\textwidth}
         \includegraphics[width=\textwidth]{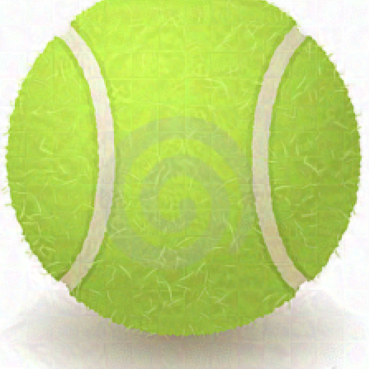}
     \end{minipage}
    \begin{minipage}[b]{0.12\textwidth}
         \centering
         \includegraphics[width=\textwidth]{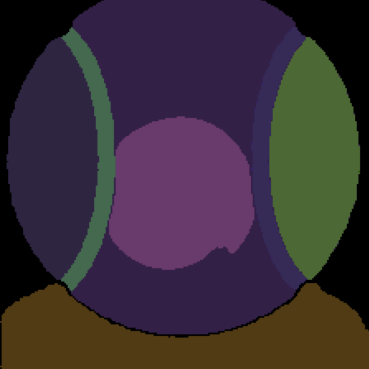}
     \end{minipage}
        \begin{minipage}[b]{0.12\textwidth}
         \centering
         \includegraphics[width=\textwidth]{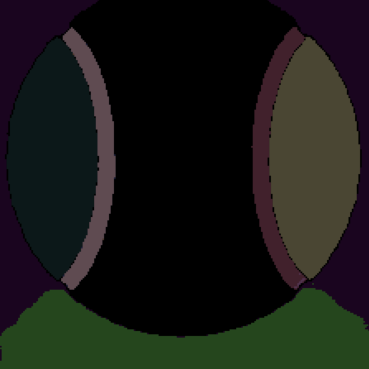}
     \end{minipage}

    \begin{minipage}[b]{0.12\textwidth}
         \includegraphics[width=\textwidth]{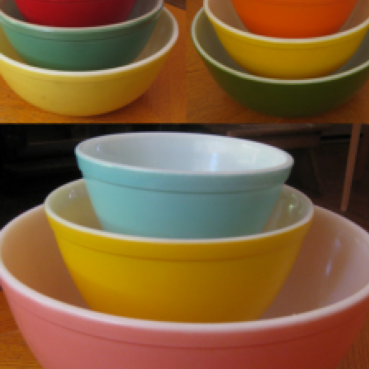}
       \subcaption{$x_{clean}$}
     \end{minipage}
    \begin{minipage}[b]{0.12\textwidth}
         \includegraphics[width=\textwidth]{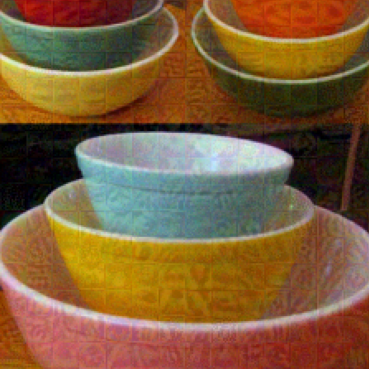}
      \subcaption{$x_{adv}$}
     \end{minipage}
    \begin{minipage}[b]{0.12\textwidth}
         \centering
         \includegraphics[width=\textwidth]{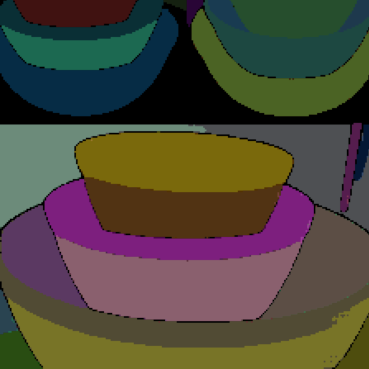}
     \subcaption{$Mask_{clean}$}
     \end{minipage}
        \begin{minipage}[b]{0.12\textwidth}
         \centering
         \includegraphics[width=\textwidth]{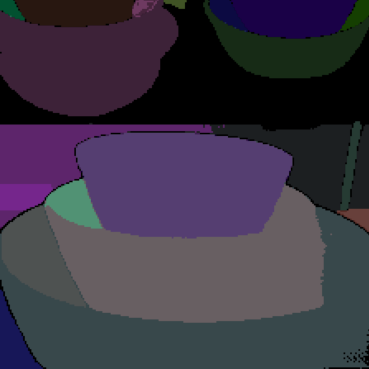}
         \subcaption{$Mask_{adv}$}
     \end{minipage}
        \caption{Masks predicted on multiple ($x$, $prompt$)  pairs in cross-task transfer task. The $x_{adv}$  is generated by a pretrained ViT model for label classification task, and then used for mask generation with SAM on multiple prompts. Compared to $Mask_{clean}$ in (c) , several masks are removed in $Mask_{adv}$ in (d)  but there are still masks remain.
        }
    \label{fig:task_transfer}
\end{figure}

\textbf{A preliminary investigation}. We conduct a cross-task transfer attack by generating $x_{adv}$ with a ViT model trained on ImageNet dataset for image classification, and then apply SAM to perform mask predictions on $x_{adv}$. As shown in Figure~\ref{fig:task_transfer}, although  $x_{adv}$  in Figure~\ref{fig:task_transfer}(b)  is generated by a classification task model, it can somehow remove several masks, comparing $Mask_{adv}$ in Figure~\ref{fig:task_transfer}(d)  to $Mask_{clean}$  in Figure~\ref{fig:task_transfer}(c) . However, it should be noted that there are still many masks remaining in $Mask_{adv}$  (Figure~\ref{fig:task_transfer}(d) ) .  More investigations on the cross-task transfer are encouraged in the future.

\section{Beyond mask removal}
\label{sec:beyond}

In the above sections, we mainly target to remove the valid masks. Here, we further investigate a more intriguing scenario to exploit adversarial example for generating any desired masks. Conceptually, it aims to create a new mask instead of just removing an old mask as discussed above.

\subsection{Mask enlargement}
\label{sec:mask_enlargement}

After investigating the mask removal attack, it is natural to ask whether it is possible to add new masks to a segmentation map. As a preliminary investigation, we first attempt to enlarge the mask area regardless of the shape or size of original mask. Following a similar procedure in  mask removal task, we replace the $Neg_{th}$ in Equation~\ref{eq:clipmse_loss} with a positive threshold $Pos_{th}$, as shown in Equation~\ref{eq:clipmse_loss_enlage}. 

\begin{equation}
\delta^{*}  =  \min_{\delta \in \mathbb{S}} || Clip(SAM(prompt, x_{clean}+ \delta) ,max=Pos_{th})  - Pos_{th}||^2
\label{eq:clipmse_loss_enlage}
\end{equation}

\begin{figure*}[!htbp]
     \centering
         \begin{minipage}[b]{0.20\textwidth}
         \includegraphics[width=\textwidth]{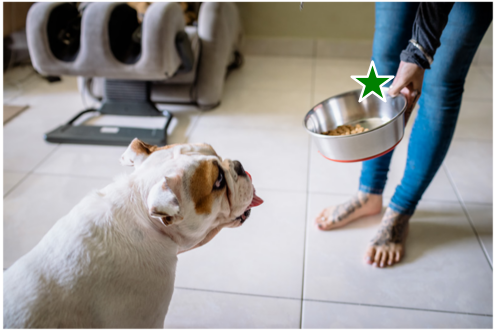}
        
     \end{minipage}
         \begin{minipage}[b]{0.20\textwidth}
         \centering
         \includegraphics[width=\textwidth]{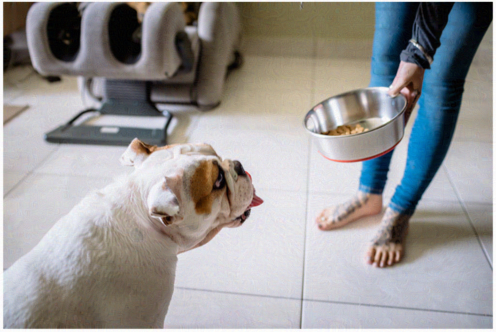}   \end{minipage}
        \begin{minipage}[b]{0.20\textwidth}
         \centering
         \includegraphics[width=\textwidth]{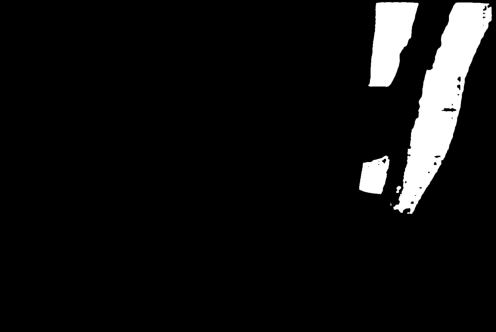}      
     \end{minipage}
     \begin{minipage}[b]{0.20\textwidth}
         \centering
         \includegraphics[width=\textwidth]{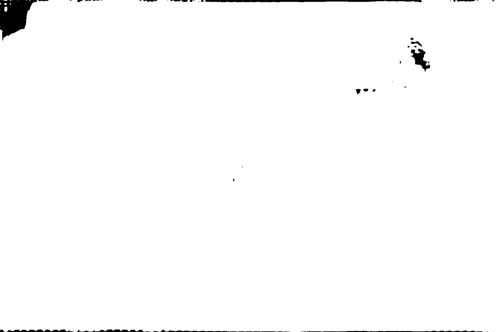}       
     \end{minipage}
      \begin{minipage}[b]{0.20\textwidth}
         \includegraphics[width=\textwidth]{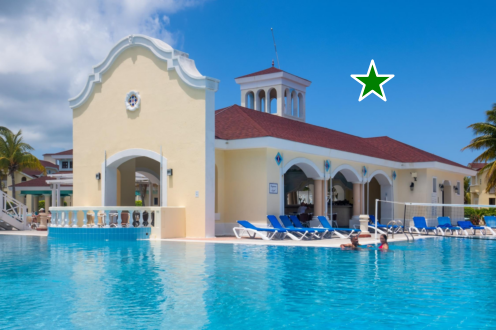}
           \subcaption{$x_{clean}$}
     \end{minipage}
         \begin{minipage}[b]{0.20\textwidth}
         \centering
         \includegraphics[width=\textwidth]{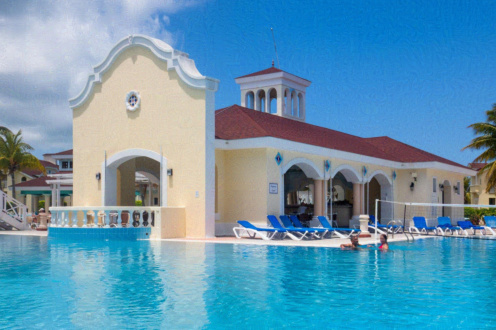}
           \subcaption{$x_{adv}$}
     \end{minipage}
        \begin{minipage}[b]{0.20\textwidth}
         \centering
         \includegraphics[width=\textwidth]{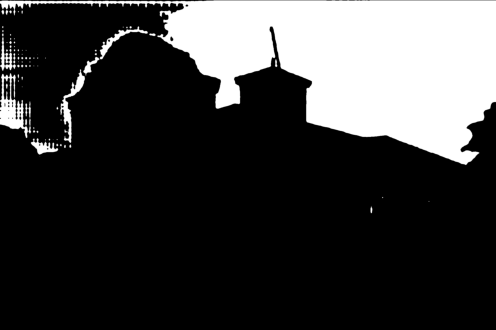}
        \subcaption{$Mask_{clean}$}
     \end{minipage}
     \begin{minipage}[b]{0.20\textwidth}
         \centering
         \includegraphics[width=\textwidth]{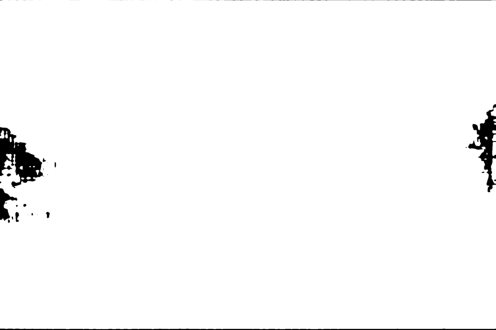}
        \subcaption{$Mask_{adv}$}
     \end{minipage}
        \caption{Results of mask enlargement attack. The $Mask_{clean}$ and $Mask_{adv}$ in (c)  (d) are generated on $x_{clean}$ and $x_{adv}$ in (a)  (b), respectively. Results show that the mask predicted by SAM could be enlarged by the adversarial attack.}
    \label{fig:enlarge_mask_whitebox}
\end{figure*}

We visualize the result of mask enlargement in Figure~\ref{fig:enlarge_mask_whitebox}.
The experimental results in Figure~\ref{fig:enlarge_mask_whitebox} show that the mask of adversarial images $Mask_{adv}$ is much larger than $Mask_{clean}$, comparing Figure~\ref{fig:enlarge_mask_whitebox}(c)  and Figure~\ref{fig:enlarge_mask_whitebox}(d) . Specifically, the entire image can be predicted as a single mask based on a randomly selected point prompt  under adversarial attack. This indicates that the adversarial attack is capable of not only shrinking the mask region but also enlarging it. In other words, under the adversarial attack, the prediction can be changed from mask to background and from background to mask. This motivates us to attack SAM for generating any desired mask.

\subsection{Mask Manipulation}
\label{sec:mask_manipulation}
Here, we manipulate the original mask on the same image. Specifically, we experiment with two use cases: mask shift and mask flipping. In the following, we highlight the main results with technical details like the loss design reported in the appendix.

\begin{figure*}[!htbp]
     \centering
           \begin{minipage}[b]{0.15\textwidth}
         \includegraphics[width=\textwidth]{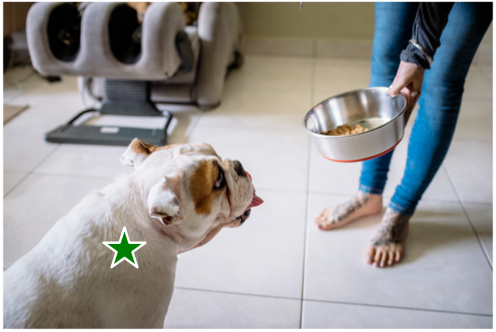}
     \end{minipage}
         \begin{minipage}[b]{0.15\textwidth}
         \centering
         \includegraphics[width=\textwidth]{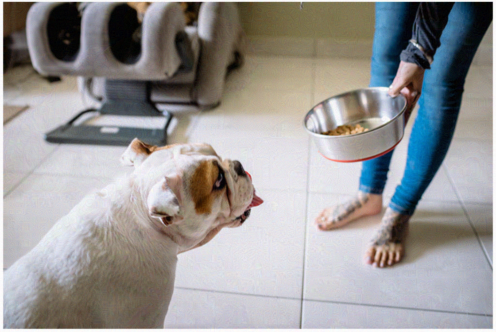}
     \end{minipage}
      \begin{minipage}[b]{0.15\textwidth}
         \centering
         \includegraphics[width=\textwidth]{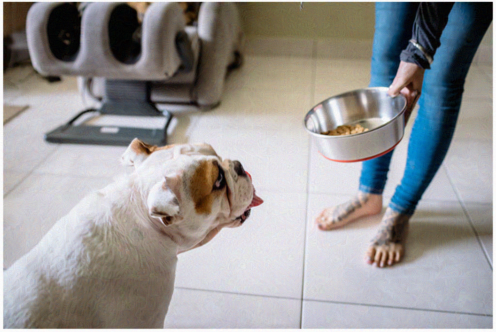}
     \end{minipage}
        \begin{minipage}[b]{0.15\textwidth}
         \centering
         \includegraphics[width=\textwidth]{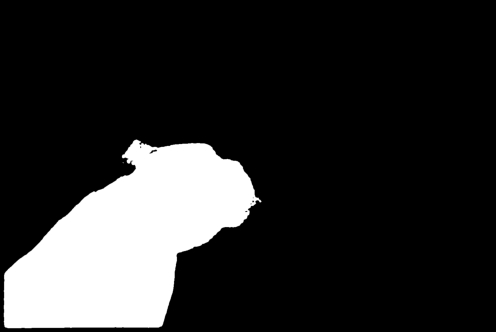}
     \end{minipage}
     \begin{minipage}[b]{0.15\textwidth}
         \centering
         \includegraphics[width=\textwidth]{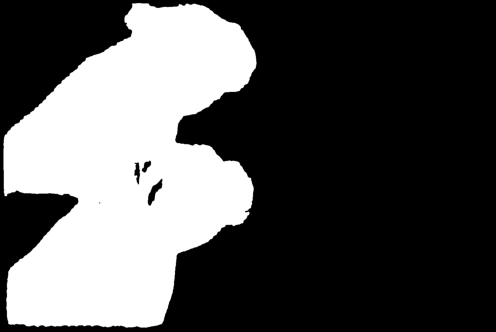}
     \end{minipage}
          \begin{minipage}[b]{0.15\textwidth}
         \centering
         \includegraphics[width=\textwidth]{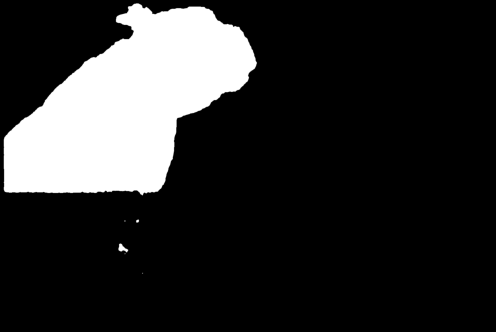}
     \end{minipage}

     \begin{minipage}[b]{0.15\textwidth}
         \includegraphics[width=\textwidth]{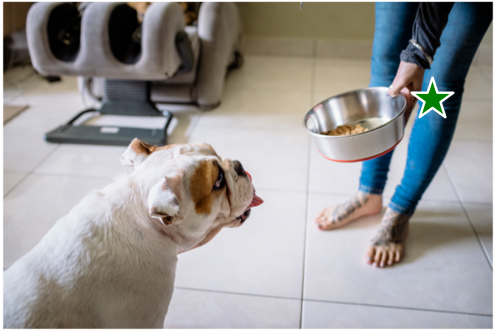}
          \subcaption{$x_{clean}$}
     \end{minipage}
         \begin{minipage}[b]{0.15\textwidth}
         \centering
         \includegraphics[width=\textwidth]{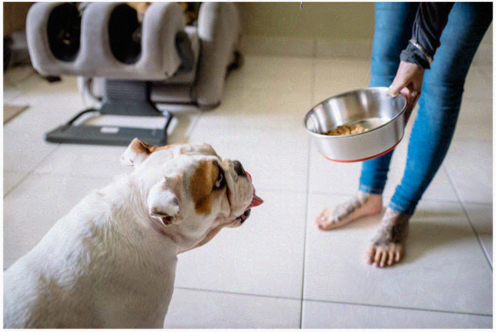}
          \subcaption{$x_{adv}^{dup}$}
     \end{minipage}
        \begin{minipage}[b]{0.15\textwidth}
         \centering
         \includegraphics[width=\textwidth]{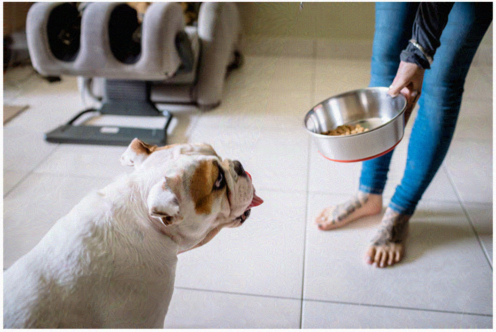}
          \subcaption{$x_{adv}^{trans}$}
     \end{minipage}
        \begin{minipage}[b]{0.15\textwidth}
         \centering
         \includegraphics[width=\textwidth]{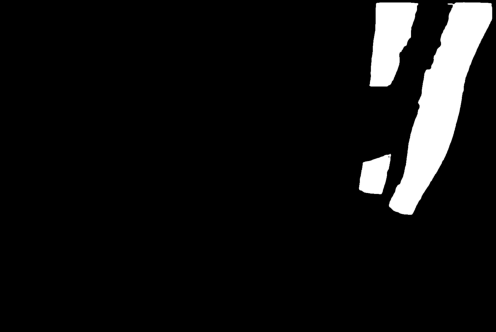}
         \subcaption{$Mask_{clean}$}
     \end{minipage}
     \begin{minipage}[b]{0.15\textwidth}
         \centering
         \includegraphics[width=\textwidth]{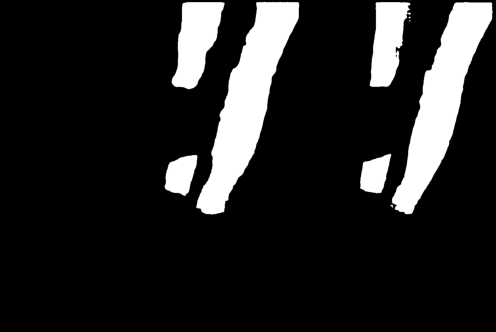}
         \subcaption{$Mask_{adv}^{dup}$}
     \end{minipage}   
      \begin{minipage}[b]{0.15\textwidth}
         \centering
         \includegraphics[width=\textwidth]{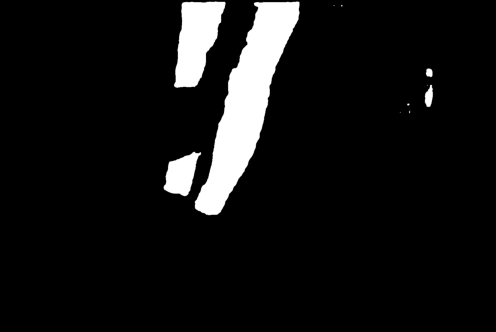}
         \subcaption{$Mask_{adv}^{trans}$}
     \end{minipage}

        \caption{Results of mask shift attack. The $Mask_{clean}$ and $Mask_{adv}$ in (d)  (e)  (f)  are generated on $x_{clean}$ and $x_{adv}$ in (a)  (b)  (c) , respectively. With adversarial attack, the original mask in (d)  could be shifted by either duplicating in (e)  or just translating in (f)  to a new position.}
    \label{fig:mask_shift}
\end{figure*}

\textbf{Mask shift.} For an existing mask in the image,  mask shift attack changes its position by either duplicating  or translating the original mask to a new location. The experimental results in Figure~\ref{fig:mask_shift} show that both duplicating  or translating the original mask could be achieved by the adversarial attack. It should be noted that translating the original mask (see Figure~\ref{fig:mask_shift}(f) )  to a new location is more difficult than duplicating it (see Figure~\ref{fig:mask_shift}(e) )  since the task in Figure~\ref{fig:mask_shift}(f)  also needs to remove its original mask. 

\begin{figure*}[!htbp]
     \centering

    \begin{minipage}[b]{0.15\textwidth}
         \includegraphics[width=\textwidth]{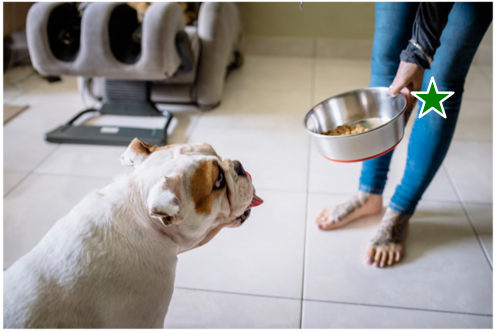}
     \end{minipage}
         \begin{minipage}[b]{0.15\textwidth}
         \centering
         \includegraphics[width=\textwidth]{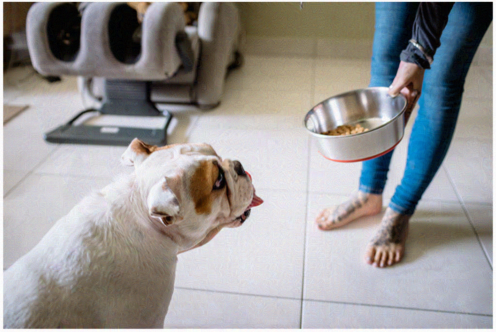}
     \end{minipage}
             \begin{minipage}[b]{0.15\textwidth}
         \centering
         \includegraphics[width=\textwidth]{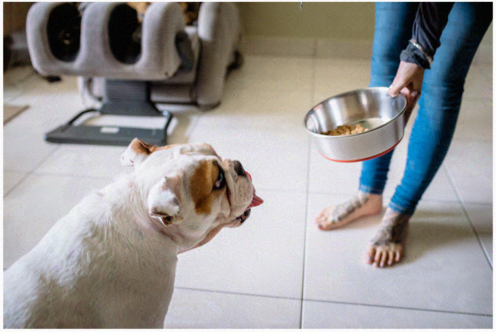}
     \end{minipage}
        \begin{minipage}[b]{0.15\textwidth}
         \centering
         \includegraphics[width=\textwidth]{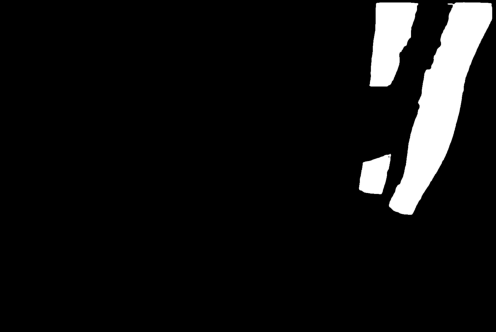}
     \end{minipage}
     \begin{minipage}[b]{0.15\textwidth}
         \centering
         \includegraphics[width=\textwidth]{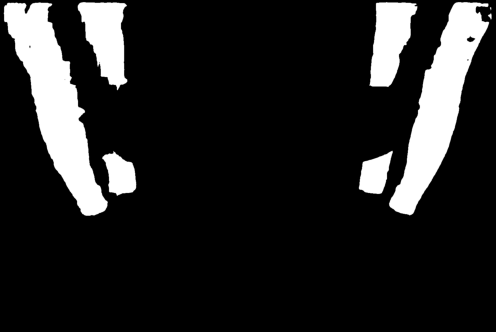}
     \end{minipage}
     \begin{minipage}[b]{0.15\textwidth}
         \centering
         \includegraphics[width=\textwidth]{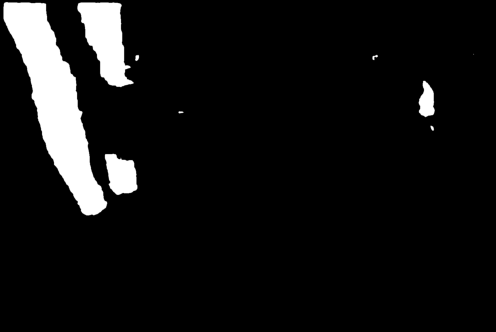}
     \end{minipage}

         \begin{minipage}[b]{0.15\textwidth}
         \includegraphics[width=\textwidth]{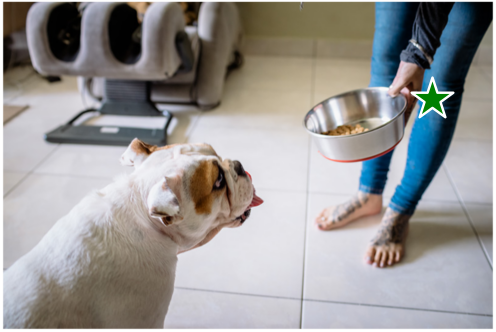}
          \subcaption{ (a)  $x_{clean}$}
     \end{minipage}
       \begin{minipage}[b]{0.15\textwidth}
         \centering
         \includegraphics[width=\textwidth]{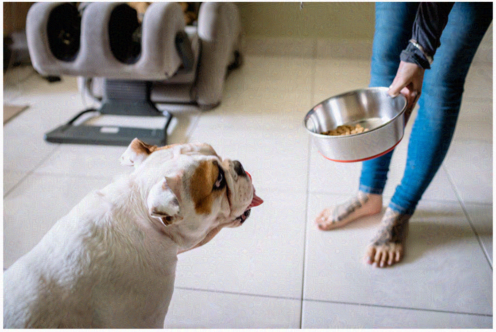}
         \subcaption{$x_{adv}^{flip}$}
     \end{minipage}
         \begin{minipage}[b]{0.15\textwidth}
         \centering
         \includegraphics[width=\textwidth]{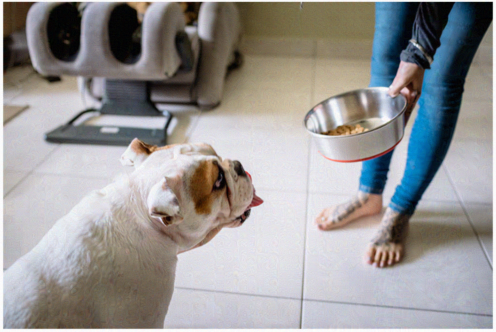}
          \subcaption{$x_{adv}^{dup}$}
     \end{minipage}
        \begin{minipage}[b]{0.15\textwidth}
         \centering
         \includegraphics[width=\textwidth]{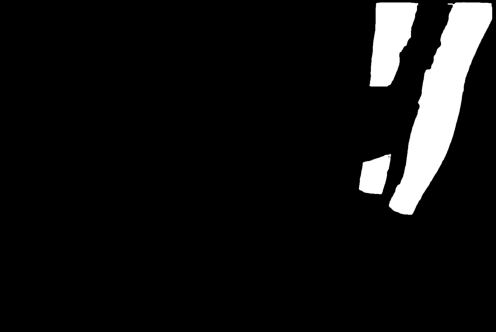}
          \subcaption{$Mask_{clean}$}
     \end{minipage}
     \begin{minipage}[b]{0.15\textwidth}
         \centering
         \includegraphics[width=\textwidth]
         {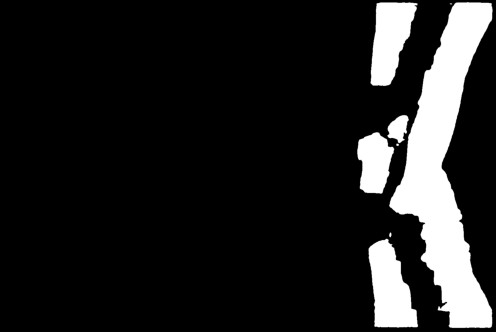}
         \subcaption{$Mask_{adv}^{dup}$}
     \end{minipage}
    \begin{minipage}[b]{0.15\textwidth}
         \centering
         \includegraphics[width=\textwidth]{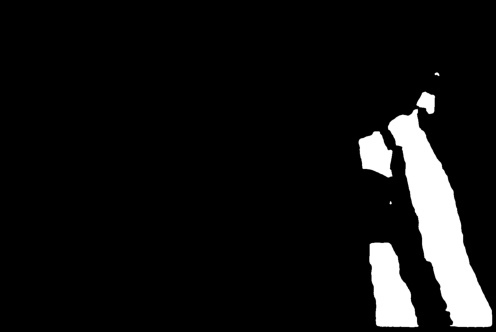}
        \subcaption{$Mask_{adv}^{flip}$}
     \end{minipage}

        \caption{Results of mask flipping attack. The $Mask_{clean}$ and $Mask_{adv}$ in (d)  (e)  (f)  are generated on $x_{clean}$ and $x_{adv}$ in (a)  (b)  (c) , respectively. With adversarial attack, the original mask in (d)  could be flipped by either duplicating in (e)  or just flipped in (f) .}
    \label{fig:mask_flip_whitebox}
\end{figure*}

\textbf{Mask flipping.} Another manipulation attempt on the original mask is to flip it. We follow the procedure of mask shift by duplicating or flipping the original mask,  as shown in  Figure~\ref{fig:mask_flip_whitebox}. The results in Figure~\ref{fig:mask_flip_whitebox} show that the original mask could be flipped either in the horizontal or vertical direction, generating new  tasks  in both duplicating (Figure~\ref{fig:mask_flip_whitebox}(e) )  and flipping Figure~\ref{fig:mask_flip_whitebox}(f) )  tasks.

\subsection{Towards generating any desired mask}
\label{sec:mask_any}

Although mask manipulation in the above Section~\ref{sec:mask_manipulation} succeeds in generating masks in different locations, they do not change the mask shape of the detected objects. Here, we experiment with a more general goal to generate any desired mask. Similar to the above investigation, we highlight the main results here and report the technical details in the appendix. We experiment with setting the target mask in three setups: (1)  a manually designed mask at a random position; (2)  a mask generated with a different point prompt  on the same image; (3)  a mask generated on a different image (with a random point prompt).

\begin{figure*}[!htbp]
     \centering

    \begin{minipage}[t]{0.15\textwidth}
         \includegraphics[width=\textwidth]{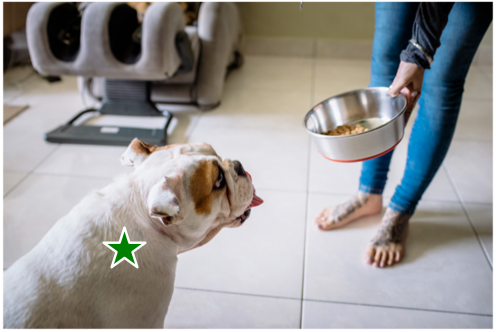}
          \subcaption{$x_{clean}$}
     \end{minipage}
      \begin{minipage}[t]{0.15\textwidth}
         \centering
         \includegraphics[width=\textwidth]{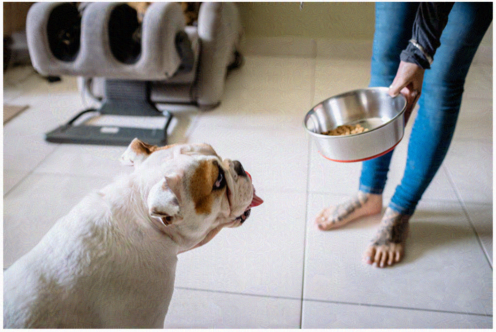}
        \subcaption{$x_{adv}^{circle}$}
     \end{minipage}
         \begin{minipage}[t]{0.15\textwidth}
         \centering
         \includegraphics[width=\textwidth]{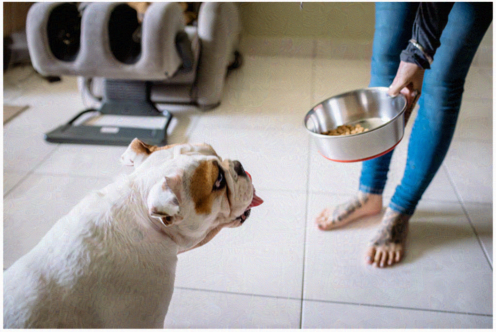}
          \subcaption{$x_{adv}^{square}$}
     \end{minipage}
        \begin{minipage}[t]{0.15\textwidth}
         \centering
         \includegraphics[width=\textwidth]{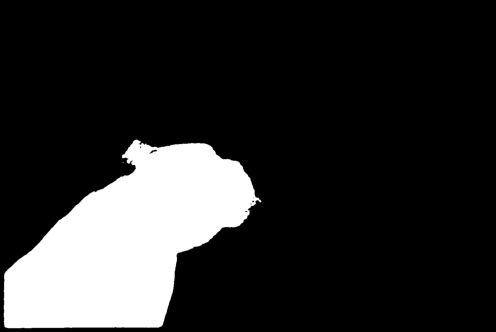}
          \subcaption{$Mask_{clean}$}
     \end{minipage}
      \begin{minipage}[t]{0.15\textwidth}
         \centering
         \includegraphics[width=\textwidth]{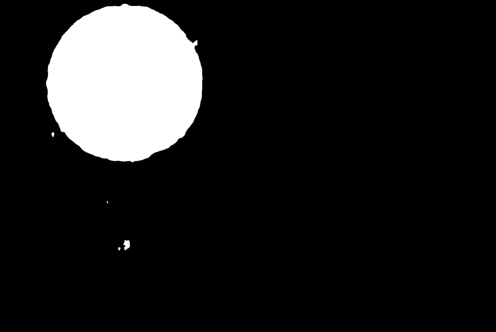}
         \subcaption{$Mask_{adv}^{circle}$}
     \end{minipage}
     \begin{minipage}[t]{0.15\textwidth}
         \centering
         \includegraphics[width=\textwidth]{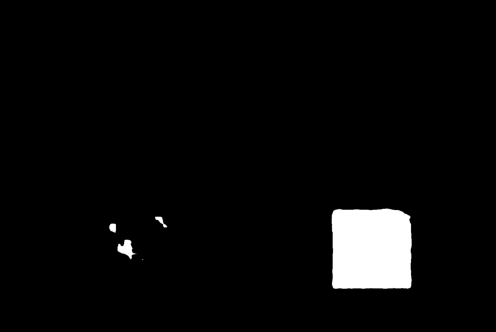}
         \subcaption{$Mask_{adv}^{square}$}
     \end{minipage}
        \caption{Towards generating any desired masks (setting 1). 
    The $Mask_{clean}$ and $Mask_{adv}$ in (d)  (e)  (f)  are generated on $x_{clean}$ and $x_{adv}$ in (a)  (b)  (c), respectively. With adversarial attack, manually designed mask in (e)  (f)  could be generated at a random position.}
    \label{fig:random_mask}
\end{figure*}

\textbf{Setting 1: a manually designed mask at a random position.} Here, we investigate whether an adversarial attack could generate manually desired masks at a random position. For not losing generality, we design masks of geometric shapes including circle and square.  Figure~\ref{fig:random_mask} shows that this goal could be achieved by setting the mask target to be a circle and square at a random position when generating  $x_{adv}$ in  Figure~\ref{fig:random_mask}(b)  and Figure~\ref{fig:random_mask}(c). Although the input prompt in Figure~\ref{fig:random_mask}(a)  expects a mask of dog as in $Mask_{clean}$ of Figure~\ref{fig:random_mask}(d), the desired circle or square masks can be obtained in $Mask_{adv}$. It is non-trivial to manually design more complex mask than circles or squares. Therefore, we further exploit the real object masks generated by the SAM as the target (reference)  masks to attack the SAM (see Setting 2 and Setting 3).

\begin{figure*}[!htbp]
     \centering
         \begin{minipage}[b]{0.20\textwidth}
         \includegraphics[width=\textwidth]{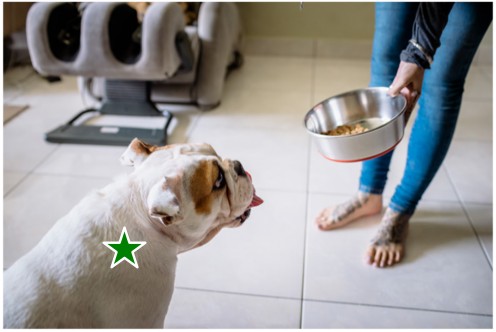}
     \end{minipage}
         \begin{minipage}[b]{0.20\textwidth}
         \centering
         \includegraphics[width=\textwidth]{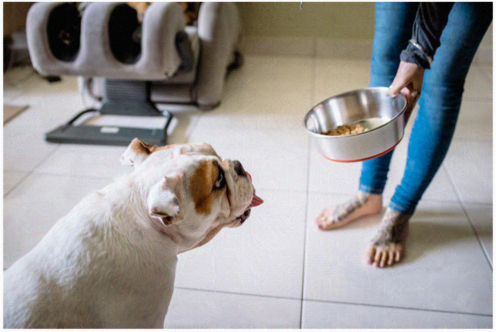}
     \end{minipage}
        \begin{minipage}[b]{0.20\textwidth}
         \centering
         \includegraphics[width=\textwidth]{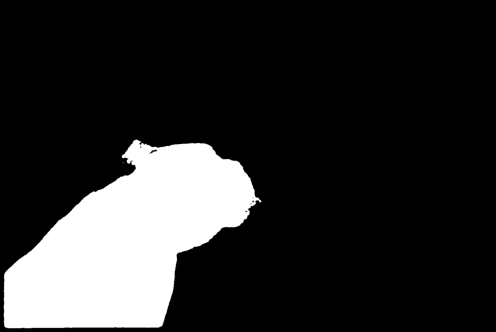}
     \end{minipage}
     \begin{minipage}[b]{0.20\textwidth}
         \centering
         \includegraphics[width=\textwidth]{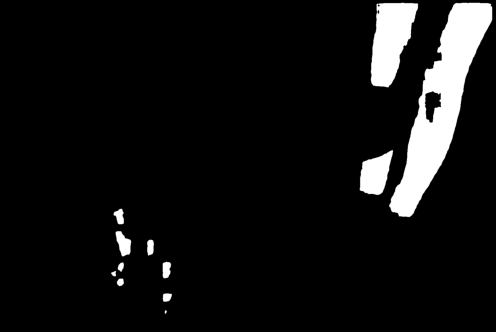}
     \end{minipage}

    \begin{minipage}[b]{0.20\textwidth}
         \includegraphics[width=\textwidth]{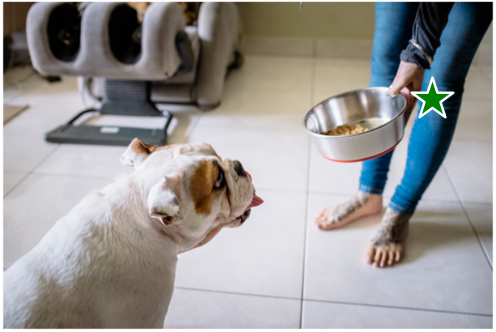}
          \subcaption{$x_{clean}$}
     \end{minipage}
         \begin{minipage}[b]{0.20\textwidth}
         \centering
         \includegraphics[width=\textwidth]{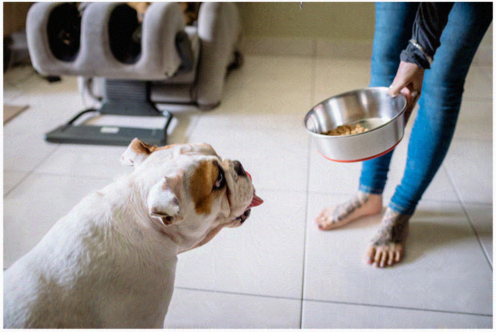}
          \subcaption{$x_{adv}$}
     \end{minipage}
        \begin{minipage}[b]{0.20\textwidth}
         \centering
         \includegraphics[width=\textwidth]{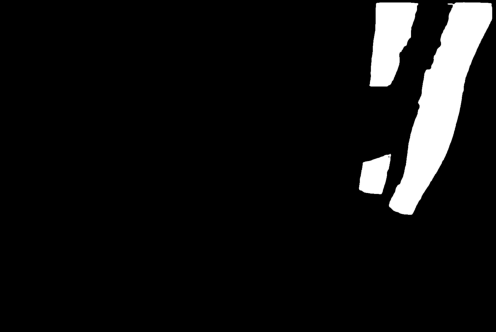}
          \subcaption{$Mask_{clean}$}
     \end{minipage}
     \begin{minipage}[b]{0.20\textwidth}
         \centering
         \includegraphics[width=\textwidth]{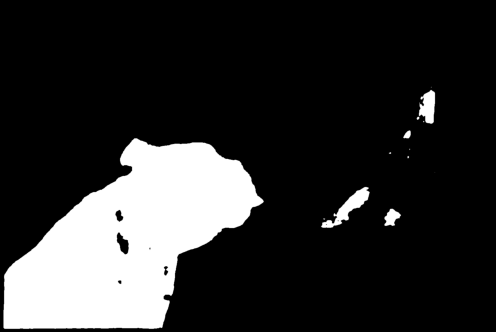}
         \subcaption{$Mask_{adv}$}
     \end{minipage}
        \caption{Towards generating any desired masks (setting 2) .
         The $Mask_{clean}$ and $Mask_{adv}$ in (c)  (d)  are generated on $x_{clean}$ and $x_{adv}$ in (a)  (b) , respectively. With adversarial attack, the mask from same image but  different point prompt in (d)  could be generated.
         }
    \label{fig:random_prompt_mask}
\end{figure*}

\textbf{Setting 2: a mask generated with a different point prompt  on the same image.} In this setup, we set the target mask to that generated on the same image with a different point prompt. The results are shown in Figure~\ref{fig:random_prompt_mask}. Let us first look at the first row in Figure~\ref{fig:random_prompt_mask}. The input point prompt $prompt_1$ in Figure~\ref{fig:random_prompt_mask}(a)  expects a mask of a dog as in $Mask_{clean}$ (see the first row of Figure~\ref{fig:random_prompt_mask}(c) ). However, the $Mask_{adv}$ predicted based on $x_{adv}$ includes human legs as in the first row  of Figure~\ref{fig:random_prompt_mask}(d). This is achieved by taking the predicted mask of another prompt $prompt_2$  as the target when generating $x_{adv}$. However, in the original dog region, the mask is not fully removed, which suggests that the attack method can be further improved. We leave this investigation to future works. A similar phenomenon can be observed in the second row of Figure~\ref{fig:random_prompt_mask}.

\begin{figure*}[!htbp]
     \centering
         \begin{minipage}[b]{0.15\textwidth}
         \includegraphics[width=\textwidth]{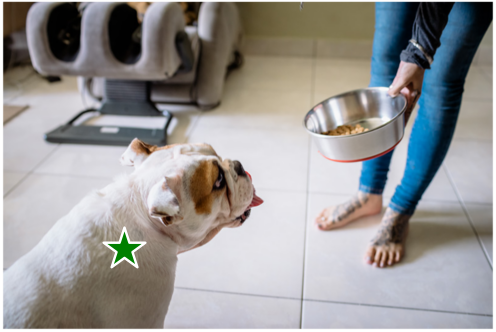}
     \end{minipage}
    \begin{minipage}[b]{0.15\textwidth}
         \includegraphics[width=\textwidth]{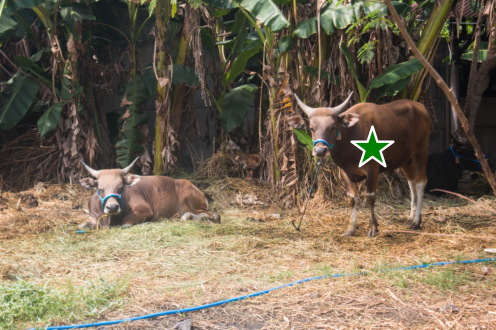}
     \end{minipage}
    \begin{minipage}[b]{0.15\textwidth}
         \centering
         \includegraphics[width=\textwidth]{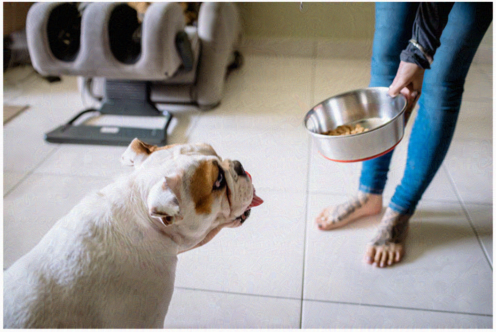}
     \end{minipage}
        \begin{minipage}[b]{0.15\textwidth}
         \centering
         \includegraphics[width=\textwidth]{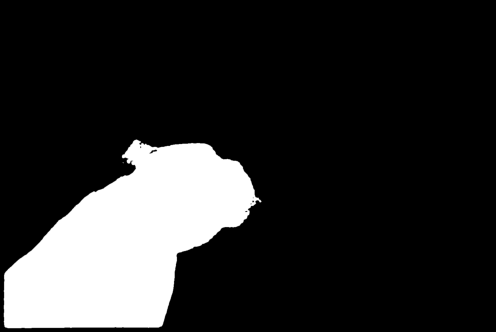}
     \end{minipage}
    \begin{minipage}[b]{0.15\textwidth}
         \includegraphics[width=\textwidth]{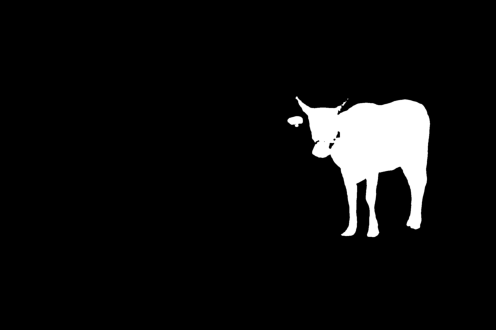}
     \end{minipage}
     \begin{minipage}[b]{0.15\textwidth}
         \centering
         \includegraphics[width=\textwidth]{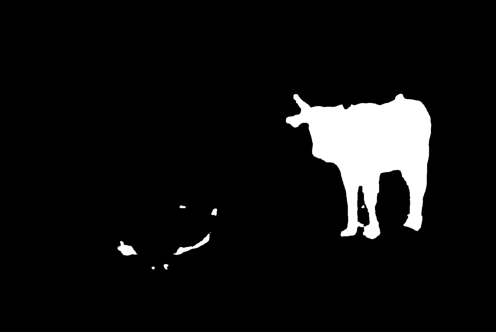}
     \end{minipage}

    \begin{minipage}[t]{0.15\textwidth}
         \includegraphics[width=\textwidth]{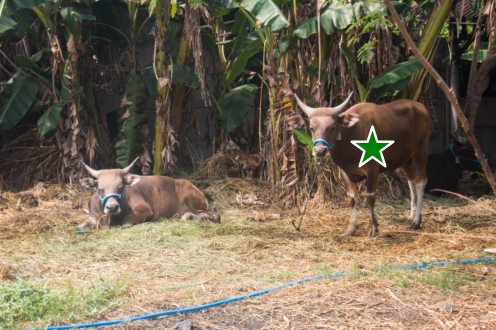}
          \subcaption{$x_{clean}$}
     \end{minipage}
         \begin{minipage}[t]{0.15\textwidth}
         \includegraphics[width=\textwidth]{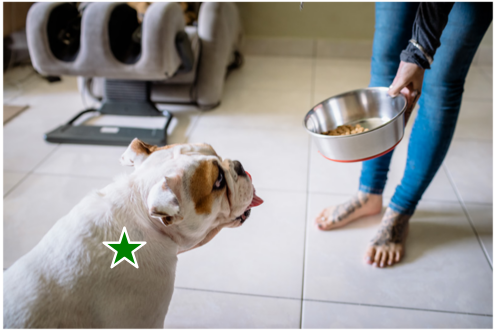}
         \subcaption{$x_{reference}$}
     \end{minipage}
         \begin{minipage}[t]{0.15\textwidth}
         \centering
         \includegraphics[width=\textwidth]{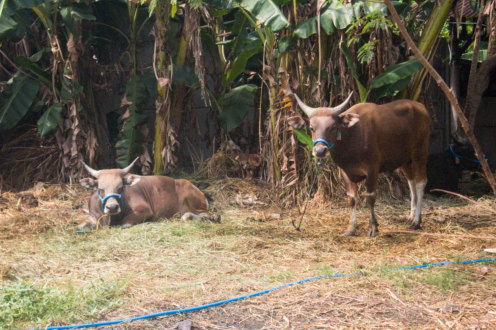}
          \subcaption{$x_{adv}$}
     \end{minipage}
        \begin{minipage}[t]{0.15\textwidth}
         \centering
         \includegraphics[width=\textwidth]{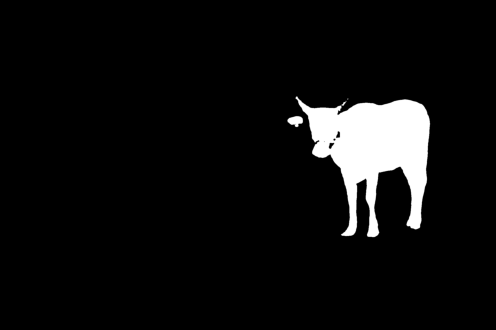}
          \subcaption{$Mask_{clean}$}
     \end{minipage}
    \begin{minipage}[t]{0.15\textwidth}
         \includegraphics[width=\textwidth]{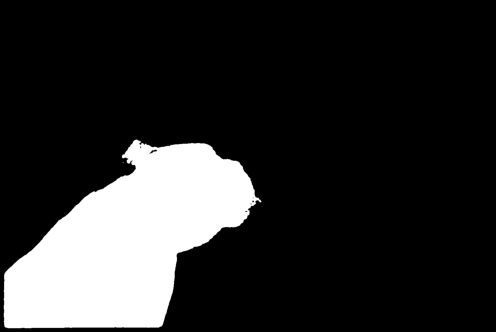}
         \subcaption{$Mask_{reference}$}
     \end{minipage}
     \begin{minipage}[t]{0.15\textwidth}
         \centering
         \includegraphics[width=\textwidth]{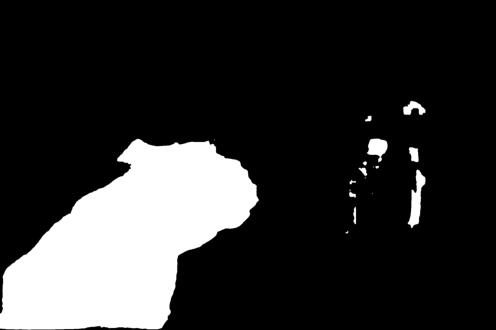}
         \subcaption{$Mask_{adv}$}
     \end{minipage}
        \caption{Towards generating any desired masks (setting 3) . The masks in (d)  (e)  (f)  are generated on images in (a)  (b)  (c) , respectively. With (b)  as a reference image and (e)  as a reference mask, the $x_{adv}$ in (c)  could predict a $Mask_{adv}$ in (f)  that is similar to $Mask_{reference}$ in (e) . }
    \label{fig:mask_from_another_image}
\end{figure*}

\textbf{Setting 3: a mask generated on a different image with a random point prompt.} We investigate this setting by two example images in Figure~\ref{fig:mask_from_another_image}. Take the first row of Figure~\ref{fig:mask_from_another_image} for example, a dog mask and a cow mask in Figure~\ref{fig:mask_from_another_image}(d)  and Figure~\ref{fig:mask_from_another_image}(e)  are predicted based on the clean images in Figure~\ref{fig:mask_from_another_image}(a)  and Figure~\ref{fig:mask_from_another_image}(b), respectively. If we take the cow mask in Figure~\ref{fig:mask_from_another_image}(e)  as $Mask_{target}$ and attack the ($x_{clean}$,$prompt$)  pair of dog image in Figure~\ref{fig:mask_from_another_image}(a) , the adversarial image $x_{adv}$ of dog image is obtained in Figure~\ref{fig:mask_from_another_image}(c) . Interestingly, a cow mask $Mask_{adv}$ is predicted in Figure~\ref{fig:mask_from_another_image}(f)  based on $x_{adv}$ of dog image in Figure~\ref{fig:mask_from_another_image}(c) . A similar observation can also be made in the second row of Figure~\ref{fig:mask_from_another_image}, predicting a dog mask in $Mask_{adv}$ of Figure~\ref{fig:mask_from_another_image}(f)  based on  $x_{adv}$ of a cow image in Figure~\ref{fig:mask_from_another_image}(c).

\section{Discussion} \label{sec:discussion}
\textbf{Attack goals: label prediction \textit{v.s.} mask prediction.} In contrast to existing works that mainly focus on attacking the model to change the label prediction, our work investigates how to attack the SAM for changing the mask prediction. Conceptually, mask removal can be interpreted as untargeted attack setting for making the adversarial mask/label prediction different from the clean mask/label prediction. Our investigation in Section~\ref{sec:beyond} to generate any desire mask is conceptually similar to the targeted attack setting. In other words, the prediction after the attack needs to align with the pre-determined target label/mask. 

\textbf{Limitations.} Overall, we demonstrate successful attack performance in various setups. However, in some challenging setups, like cross-task attacks, the success is only partial, which is somewhat expected considering the task discrepancy between label prediction and mask prediction. For the case of generating any desired task, we find that the generated mask is not always perfect with some small noise masks in the unintended region. Future works can investigate how to improve the attack performance by removing those noise masks. Moreover, our choice of reference mask is still limited, and thus more diverse mask types can be explored in future works.

\section{Conclusion}
Our work conducts the first yet comprehensive investigation on how to attack SAM with adversarial examples. In the full white-box setting, SAM is found to be vulnerable with successful removal of the original mask. When the  prompt  is not given, the attacker can enhance the cross-prompt transferability by attacking multiple masks with different prompts when generating the adversarial image. We also experiment with the cross-task transferability and find that the adversarial examples generated to attack the semantic label prediction can also be used to attack the mask prediction to some extent. Beyond the basic attack goal of mask removal, we also attempt to generate any desired task with an overall satisfactory attack performance. It is not our intention to find the strongest method to attack SAM. Instead, we focus on adapting the common attack methods from attacking label prediction to attacking mask prediction to investigate whether SAM is robust against the attack of adversarial examples. Our finding that SAM is vulnerable to adversarial examples suggests a need to examine the safety concerns of deploying SAM in safety-critical scenarios. Future works are expected to identify stronger attack methods as well as enhance the robustness of SAM.

\bibliographystyle{plain}
\bibliography{bib_mixed,bib_local,bib_sam}

\end{document}